\documentclass[10pt,twocolumn,letterpaper]{article}

\usepackage{cvpr}              

\usepackage[dvipsnames]{xcolor}

\usepackage{graphicx}
\usepackage{amsmath}
\usepackage{amssymb}
\usepackage{booktabs}
\usepackage[accsupp]{axessibility}

\usepackage{multirow}
\usepackage{algorithm}
\usepackage{algorithmic}
\usepackage[T1]{fontenc}
\usepackage{multicol}
\usepackage{stfloats}
\usepackage{xcolor}

\usepackage{mathtools}

\usepackage{makecell}
\usepackage{scalerel}
\usepackage{array}
\usepackage{dsfont}
\usepackage{url}
\usepackage{pifont}
\usepackage{adjustbox}

\newcolumntype{C}[1]{>{\centering\let\newline\\\arraybackslash\hspace{0pt}}p{#1}}
\newcolumntype{R}[1]{>{\raggedleft\let\newline\\\arraybackslash\hspace{0pt}}p{#1}}
\newcolumntype{L}[1]{>{\raggedright\let\newline\\\arraybackslash\hspace{0pt}}p{#1}}
\newcolumntype{M}[1]{>{\centering\let\newline\\\arraybackslash\hspace{0pt}}m{#1}}

\newcommand\Tstrut{\rule{-3pt}{2.6ex}}       
\newcommand\Bstrut{\rule[-0.9ex]{-3pt}{0pt}} 
\newcommand{\TBstrut}{\rule{-3pt}{2.6ex} \rule[-0.9ex]{-2pt}{0pt}}  

\newcommand\mydots{\hbox to 1em{.\hss.\hss.}}

\definecolor{myblue}{RGB}{0, 250, 0}
\definecolor{mypink}{RGB}{237, 2, 140}
\definecolor{green1}{RGB}{148, 193, 117}
\definecolor{green2}{RGB}{90, 138, 57}

\newcolumntype{D}[2]{%
    >{\adjustbox{angle=#1,lap=\width-(#2)}\bgroup}%
    l%
    <{\egroup}%
}
\newcommand*\rott{\multicolumn{1}{D{45}{1em}}}

\definecolor{cvprblue}{rgb}{0.21,0.49,0.74}
\usepackage[pagebackref,breaklinks,colorlinks,citecolor=cvprblue]{hyperref}

\def\paperID{1815} 
\def\confName{CVPR}
\def\confYear{2024}

\title{KPConvX: Modernizing Kernel Point Convolution with Kernel Attention}

\author{Hugues Thomas$^\dagger$, Yao-Hung Hubert Tsai$^\dagger$, Timothy D. Barfoot$^\ddagger$, Jian Zhang$^\dagger$ \\
  $^\dagger$Apple, $^\ddagger$University of Toronto \\ 
  {\small\url{https://github.com/apple/ml-kpconvx} }
}

\begin{document}
\maketitle
\begin{abstract}

In the field of deep point cloud understanding, KPConv is a unique architecture that uses kernel points to locate convolutional weights in space, instead of relying on Multi-Layer Perceptron (MLP) encodings. While it initially achieved success, it has since been surpassed by recent MLP networks that employ updated designs and training strategies. Building upon the kernel point principle, we present two novel designs: KPConvD (depthwise KPConv), a lighter design that enables the use of deeper architectures, and KPConvX, an innovative design that scales the depthwise convolutional weights of KPConvD with kernel attention values. Using KPConvX with a modern architecture and training strategy, we are able to outperform current state-of-the-art approaches on the ScanObjectNN, Scannetv2, and S3DIS datasets. We validate our design choices through ablation studies and release our code and models.

\end{abstract}    

\section{Introduction}
\label{sec:intro}

The field of 3D point cloud understanding has experienced significant growth in the past decade. This growth can be attributed to the availability of advanced 3D sensors and the increasing use of deep learning in various research domains. The evolution of this field has known different phases. Until 2017, most of the proposed approaches relied on projection in images or 3D grids \cite{maturana2015voxnet, su2015multi, riegler2017octnet, wang2017cnn, boulch2017unstructured, lawin2017deep}. However, after 2017, point-based methods gained dominance \cite{qi2017pointnet, qi2017pointnet++, li2018so, li2018pointcnn, hua2018pointwise, atzmon2018point, thomas2019kpconv, wu2019pointconv, xu_paconv_2021}, leading to a rapid expansion of the field. More recently, 3D point cloud understanding followed the trend of attention and transformer networks, which are widely adopted in other deep learning fields \cite{zhao_point_2021, lu_3dpct_2022, yu_point-bert_2022, park_fast_2022, lai_stratified_2022, wu2022point}. This diverse range of approaches is a response to the challenge posed by the unstructured and continuous nature of point clouds. In this dynamic field, we explore the potential of Kernel Point Convolution (KPConv) \cite{thomas2019kpconv}, one of the most successful point-based methods, when enhanced with modern techniques and attention mechanisms.

\begin{figure}[t]
    \centering
    \adjincludegraphics[width=0.99\columnwidth,trim={{.08\width} {.04\width} {.04\width} {.02\width}},clip]{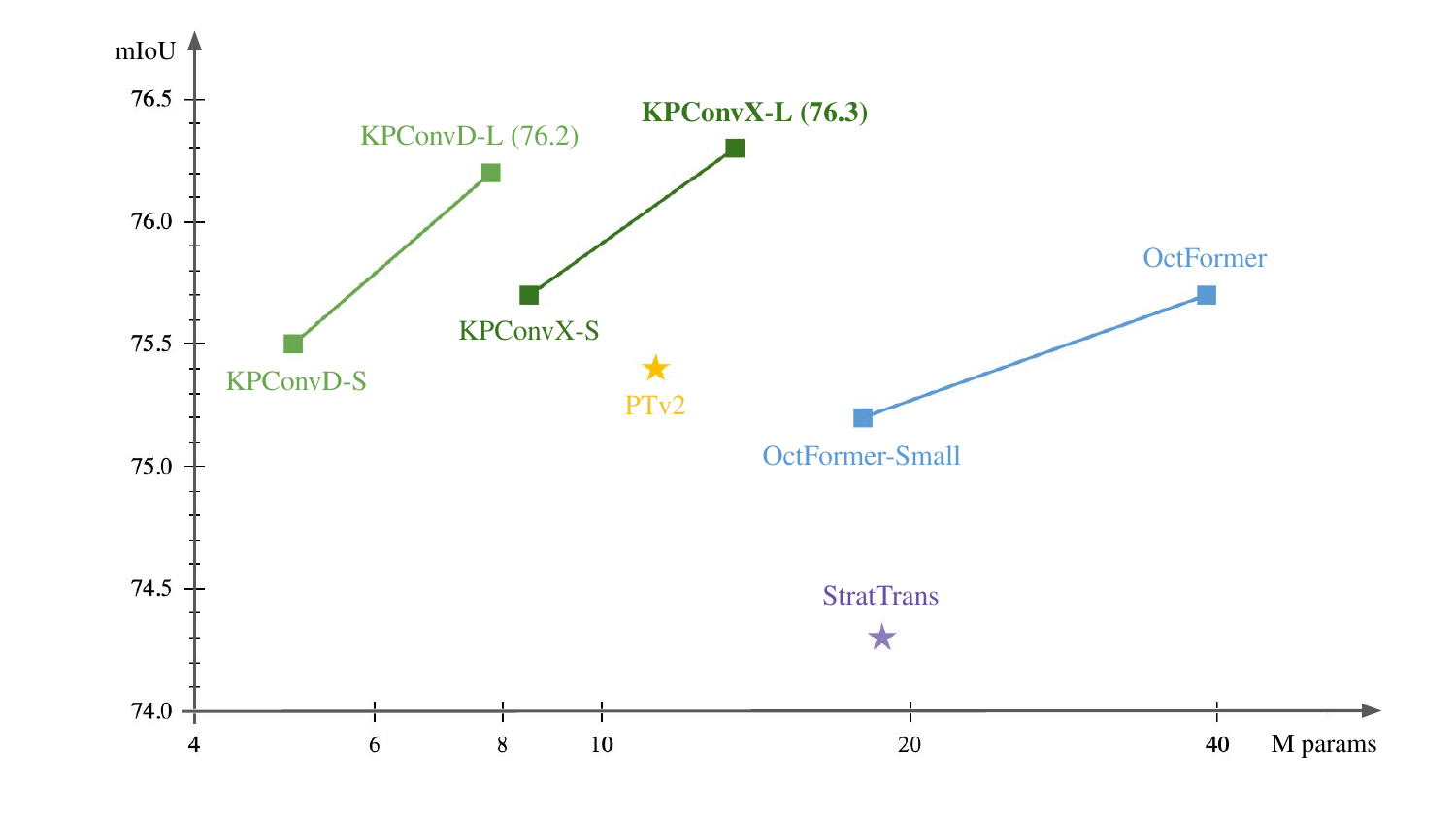}
    \caption{KPConvD and KPConvX using small (S) and large (L) architectures outperform other state-of-the-art architectures on ScanNetv2 dataset using a relatively small number of parameters.}
    \label{fig:intro_scores}
\end{figure}

\begin{figure*}[t!]
    \centering

    \adjincludegraphics[width=0.99\textwidth,trim={{.008\width} {.15\height} {.008\width} {.05\height}},clip]{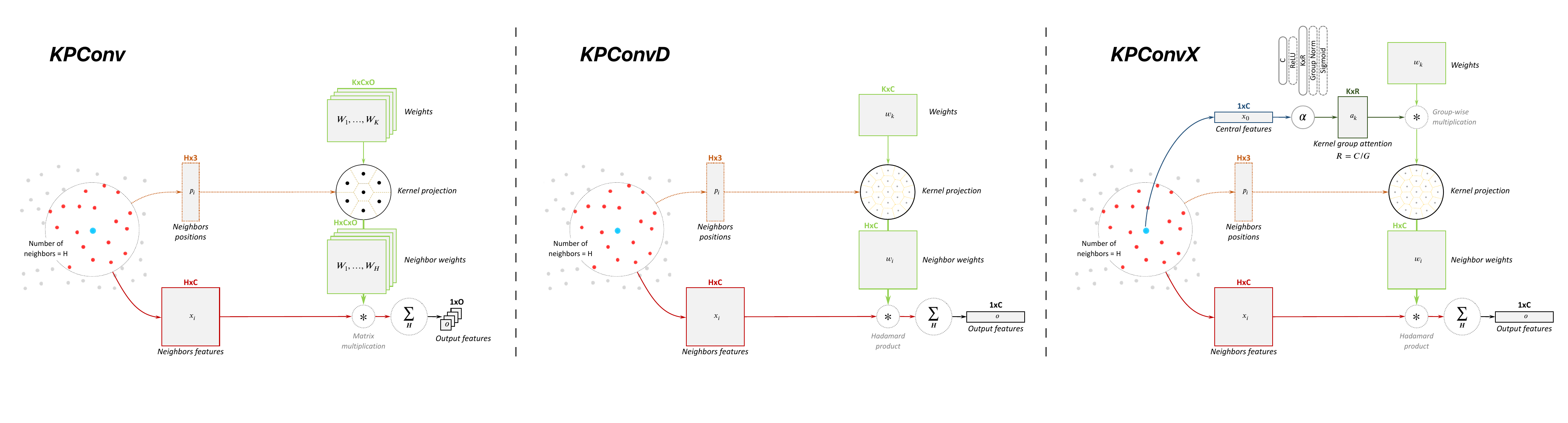}
    
    \caption{Illustration of our new designs compared to the original KPConv operator. KPConvD adopts a lighter depthwise design and KPConvX includes kernel attention.}
    \label{fig:intro}
    \vspace{-3ex}
\end{figure*}

KPConv proposed a convolution design that utilizes kernel points to store network weights in specific positions within the space. In this study, we revisit this design and introduce a new, lighter operator called KPConvD. Additionally, we enhance the operator by incorporating "geometric attention", a novel self-attention mechanism applied to the kernel points, resulting in KPConvX. Apart from the convolution operator itself, we also reconsider the network architectures built using this operator. Due to its lightweight nature, we are able to construct deeper architectures. Overall, we observe that our networks are faster than the original KPConv and achieve state-of-the-art performance.

In \cref{sec:related}, we will provide a comprehensive overview of the different approaches to deep point cloud understanding. Our focus will be on point-based methods, and we will compare KPConvX to other similar designs. \cref{sec:design} is dedicated to the definition of our convolution design. As depicted in \cref{fig:intro}, we utilize kernel points similar to KPConv, but we introduce two major changes: depthwise weights, which combine neighbors' features using the Hadamard product, and a new concept called \textbf{kernel attention} that scales the kernel weights based on the current input features. This approach, which defines attention geometrically rather than topologically, is analogous to image involution \cite{li2021involution} and is tailored for kernel points, which have a fixed order and spatial disposition, as opposed to neighbor points. Additionally, three other significant modifications are not illustrated in \cref{fig:intro}. When projecting kernel weights onto neighbors, we simply use the nearest kernel point instead of summing the influences of all kernel points. This nearest-kernel implementation allows for a larger number of kernel points without significantly increasing computations. Furthermore, we define the positions of kernel points using two shells, as proposed in \cite{li2021spnet}. Lastly, we also limit the number of neighbors to a fixed value, enabling faster computations.

\begin{figure*}[t!]
    \centering
    \includegraphics[width=0.98\textwidth, keepaspectratio=true]{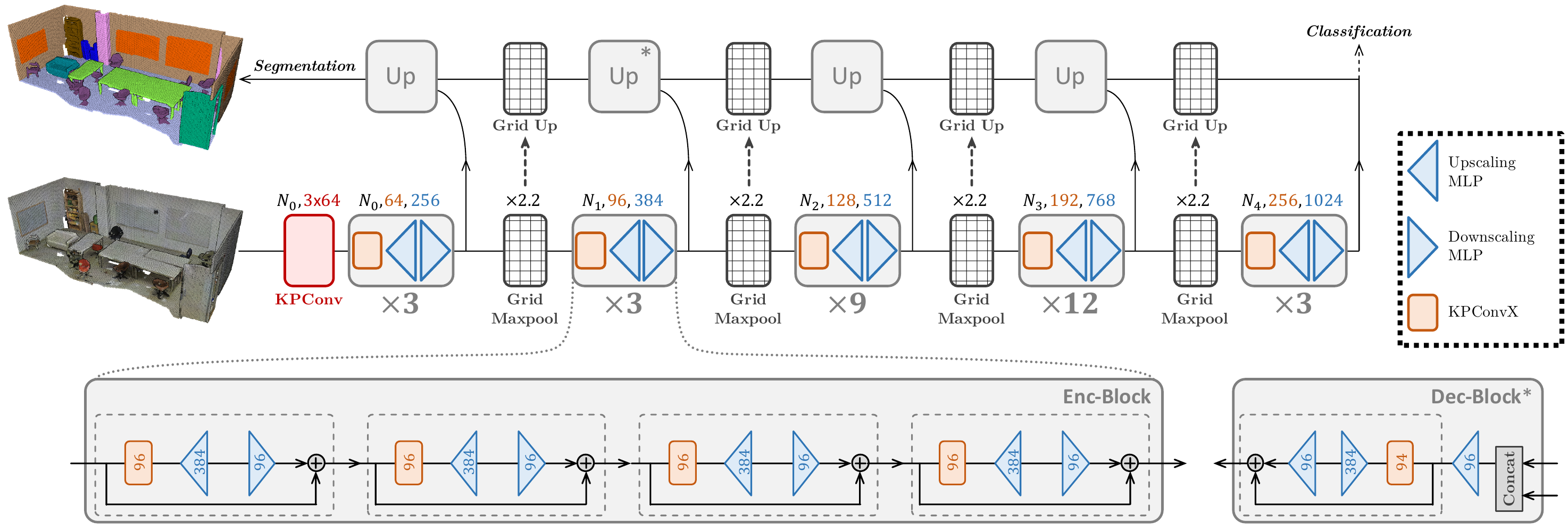}
    \caption{Illustration of our network architecture KPConvX-L. It can be used for semantic segmentation or shape classification. It has a total of $44$ encoder blocks, plus one stem KPConv and 4 decoder blocks. We use inverted bottleneck blocks and grid subsampling from one layer to the next.}
    \label{fig:network}
    \vspace{-3ex}
\end{figure*}

This lightweight operator allows for much deeper networks. As suggested in other studies that modernize deep networks \cite{liu2022convnet, qian_pointnext_2022, lai_stratified_2022}, we increase the depth (number of layers) and width (number of channels) of our networks and use a KPConv stem as the first layer. We also employ partition-based pooling for downsampling layers and incorporate convolution blocks in the decoder upsampling, as described in \cite{wu2022point}. Finally, we adopt a scaling factor of $\sqrt{2}$ (instead of 2) for the number of channels from one layer to the next. The overall network architecture is illustrated in \cref{fig:network}.

The impact of these choices is evaluated in our experiments in \cref{sec:exp}. Ablation studies demonstrate that our contributions result in better performance. Our networks outperform the state of the art on object classification and semantic segmentation tasks, ranking first on the ScanObjectNN \cite{uy2019revisiting}, S3DIS-Area5 \cite{armeni20163d}, and Scannetv2 \cite{dai2017scannet} benchmarks. \cref{fig:intro_scores} highlights the superiority of our approach over other competing methods on Scannetv2. KPConvX-L is our best-performing network, while KPConvD-L offers a great compromise between performance and efficiency.

In short, our contributions are as follows:
\begin{itemize}
    \itemsep 0ex 
    \item We introduce \textbf{KPConvD}, a novel kernel point depthwise convolution operator.
    \item We introduce \textbf{KPConvX}, a version of KPConvD augmented with kernel point attention.
    \item We design modern \textbf{architectures} showing state-of-the-art performances using our operators.
\end{itemize}

We also aim to provide a comprehensive open-source code, allowing the reproduction of our results and the use of KPConvX as a building block for other networks and tasks.

\section{Related Work}
\label{sec:related}


\noindent \textbf{Deep point cloud understanding}: As explained in the introduction, point clouds are fundamentally different from images. They are unstructured yet permutation invariant, and their data is continuous but sparse. Due to this complex nature, numerous approaches have been proposed for point cloud understanding.

One of the early ideas for processing point clouds with deep networks was to project points onto a 3D grid \cite{maturana2015voxnet, su2015multi}. The use of sparse structures like hash-maps or octrees allowed these approaches to scale to larger grids and achieve improved performance \cite{riegler2017octnet, wang2017cnn, graham20183d, su2018splatnet, choy20194d, chen2023largekernel3d}. Alternatively, points can be projected onto 2D image planes and processed with image networks \cite{su2015multi, boulch2017unstructured, lawin2017deep, tatarchenko2018tangent, lang2019pointpillars}. However, quantizing irregular point clouds onto regular grids inevitably results in data loss, and grids lack flexibility.

\noindent \textbf{Point-wise methods}: To address these problems, it is possible to directly extract features from the point clouds instead of using projections \cite{qi2017pointnet}. Typically, features are mixed using linear layers, and geometric patterns are extracted using a local operator. The definition of this local operator is crucial and often the main distinguishing factor between different methods. Many approaches utilize Multi-Layer Perceptrons (MLP) as feature extractors, operating directly on point coordinates \cite{qi2017pointnet++, li2018so, li2018pointcnn, hua2018pointwise, zhao2019pointweb, wu2019pointconv, xu_paconv_2021, ma_rethinking_2022, wijaya_advanced_2022, choe_pointmixer_2022, qian_pointnext_2022, deng2023pointvector}. MLP extractors can be viewed as combinations of hyperplanes, and with a sufficient number of parameters (planes), they can represent any geometric pattern. However, complex patterns require a large number of parameters.

\noindent \textbf{3D transformers}: Transformer-based networks utilize attention as their primary feature extraction mechanism. Whether they employ voxel grouping \cite{park_fast_2022, lai_stratified_2022, wang2023octformer} or local point neighborhoods \cite{zhao_point_2021, lu_3dpct_2022, yu_point-bert_2022, wu2022pointconvformer, wu2022point, park2023self}, they always employ MLP to encode point features and geometric patterns. As a result, they face the same limitation as MLP-based convolutions. Additionally, their attention mechanism is based on the combination of neighbors' keys with central queries, whereas we adopt a simpler central-generated kernel attention similar to \cite{li2021involution}.

\noindent \textbf{Geometric convolutions}: Eventually, local feature extractors can be defined geometrically \cite{hua2018pointwise, atzmon2018point, thomas2019kpconv, li2021spnet}. This reduces the complexity of the convolution operator without losing data through quantization. We use this approach and incorporate an attention mechanism to the convolution weights, similar to the modulations used in the deformable version of KPConv \cite{thomas2019kpconv}. However, instead of a single modulation per kernel point, we use group modulations as described in \cite{li2021involution}.

\section{Our Designs}
\label{sec:design}

This section describes our novel convolution designs called KPConvD and KPConvX, illustrated in \cref{fig:intro}. We first discuss the modifications made to the convolution itself, followed by an explanation of the kernel attention mechanism that we added to the operator. Finally, we provide an overview of our new network architectures (\cref{fig:network}) and training strategies.

%
%
%
%

\subsection{KPConvD: Depthwise Kernel Point Convolution}

In essence, our KPConvD operator can be understood as a depthwise version of KPConv. However, aside from some hyperparameter adjustments, the core implementation has been optimized for efficiency. We provide a detailed description of the modifications, using notations similar to those used in the original KPConv paper, where $K$ represents the number of kernel points and $H$ represents the number of neighbors:
\begin{itemize}
    \itemsep 0ex 
    \item input points: $x_i \in \mathbb{R}^3$ with $i<H$
    \item input features: $f_i \in \mathbb{R}^{C}$ with $i<H$
    \item kernel points: $\widetilde{x}_k \in \mathbb{R}^3$ with $k<K$
    \item kernel weights: $W_k \in \mathbb{R}^{C \times O}$ with $k<K$
\end{itemize}
\noindent Therefore, a standard KPConv can be written as

\begin{equation} \label{eq:1}
    (\mathcal{F}*g)(x) = \sum_{x_i \in \mathcal{N}_x} \sum_{k<K} h\left(x_i-x,  \widetilde{x}_k\right) W_k f_i \: ,
\end{equation}

\noindent where $\mathcal{N}_x$ is a radius neighborhood with $r \in \mathbb{R}$, and $h$ is the influence (or correlation) function. Note that in fact, the implementation of KPConv uses a radius neighborhood truncated by a maximum number of neighbors for efficiency. It is therefore equivalent to a KNN neighborhood, where all points further than $r$ are ignored, which is how we implement it in our work. We thus have a fixed number of neighbors, named $H$. We use the same influence function as KPConv:

\begin{equation} \label{eq:2}
    h_{ik} = \max\left(0, 1 - \frac{\left\Vert x_i-x - \widetilde{x}_k \right\Vert}{\sigma}\right) \: ,
\end{equation}

\noindent
where $\sigma$ is the influence distance of the kernel points. We use the notation $h_{ik}$ in place of $h\left(x_i-x,  \widetilde{x}_k\right)$ for convenience, but it also represents an optimization in our implementation. All the blocks of the same layer can share the same kernel points and therefore share the same influences. We thus compute the matrix of influences $(h_{ik}) \in \mathbb{R}^{H \times K}$ only once per layer and share the values $h_{ik}$ for the rest of the layer

With these new notations, the depthwise equivalent of KPConv can be defined as

\begin{equation} \label{eq:3}
    (\mathcal{F}*g)(x) = \sum_{i<H} \sum_{k<K} h_{ik} \, w_k \odot f_i \: ,
\end{equation}

\noindent
where $w_k \in \mathbb{R}^{C}$ is a single vector of weights instead of a full matrix and $\odot$ is the Hadamard product. We observe that the influence matrices usually contain a majority of zeros. Indeed, the kernel points only apply their weights to the neighbors that are in their area of influence. However, in this definition, the influence of all kernel points is summed for each neighbor, which is a waste of computations. We reduce the number of operations by choosing a single kernel point for each neighbor instead:

\begin{equation} \label{eq:4}
    (\mathcal{F}*g)(x) = \sum_{i<H} h_{ik^*} \, w_{k^*} \odot f_i \: ,
\end{equation}

\noindent
where $k^*$ is the index of the nearest kernel point to the neighbor $i$. With this nearest-kernel implementation, we are able to increase the number of kernel points and thus the descriptive power of the convolution without affecting the number of operations.

\begin{figure}[t]
    \centering
    \includegraphics[width=0.99\columnwidth, keepaspectratio=true]{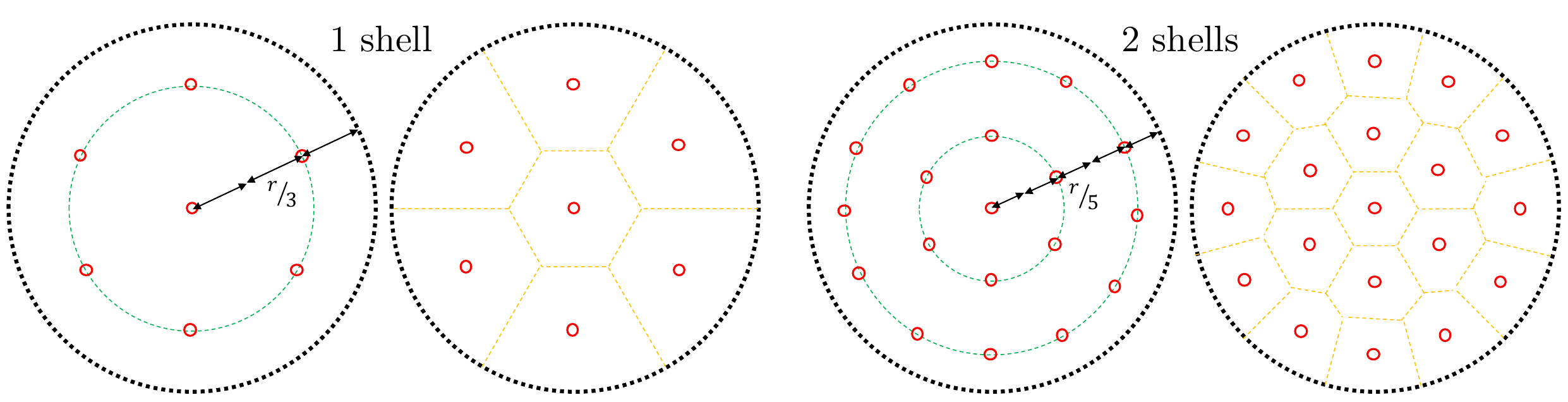}
    \caption{Illustration of kernel dispositions in 2D with one or two shells. Red circles are kernel points. The shells highlighted in green are placed regularly along the radius (left). The nearest-kernel area of each kernel point is shown in yellow (right).}
    \label{fig:shells}
    \vspace{-2ex}
\end{figure}

Because of this ability to increase the number of kernel points, we adopt a shell definition of the kernel point positions as proposed in \cite{li2021spnet}, but keep the convolution definition defined in \cref{eq:4}. The kernel point locations are found with a similar optimization process to the one used in KPConv, but with an equality constraint added for the radius of the points of each shell. More details can be found in the supplementary material. We note the number of kernel points as a list $\left[1, N_1, ..., N_s\right]$, where $s$ is the number of shells. In the rest of the paper, we use a two-shell kernel disposition $\left[1, 14, 28\right]$. Examples of 2D kernel points' disposition and their corresponding nearest-kernel areas are shown in \cref{fig:shells}.

Eventually, we choose a radius of $r = 2.1$ times the subsampling grid size, slightly smaller than the original KPConv. This value was found empirically (see additional experiments in the supplementary material). The chosen radius does not affect memory consumption as the number of neighbors is fixed. Because of our nearest-kernel implementation, the area of influence of each kernel point does not overlap, and we use a large influence radius $\sigma = r$. We note that this nearest-kernel design was proposed in the thesis \cite{thomas2019learning}, but never implemented.

%
%
%
%

\subsection{KPConvX: Adding Kernel Attention}

Local self-attention, as it is commonly used in transformers for point clouds \cite{zhao_point_2021, wu2022point}, is defined with respect to the neighbor features. It is a topological operator, which combines features without extracting geometric information. This is why most of the self-attention designs for point clouds incorporate some kind of geometric encoding.

\begin{figure}[b]
    \centering
    \includegraphics[width=0.99\columnwidth, keepaspectratio=true]{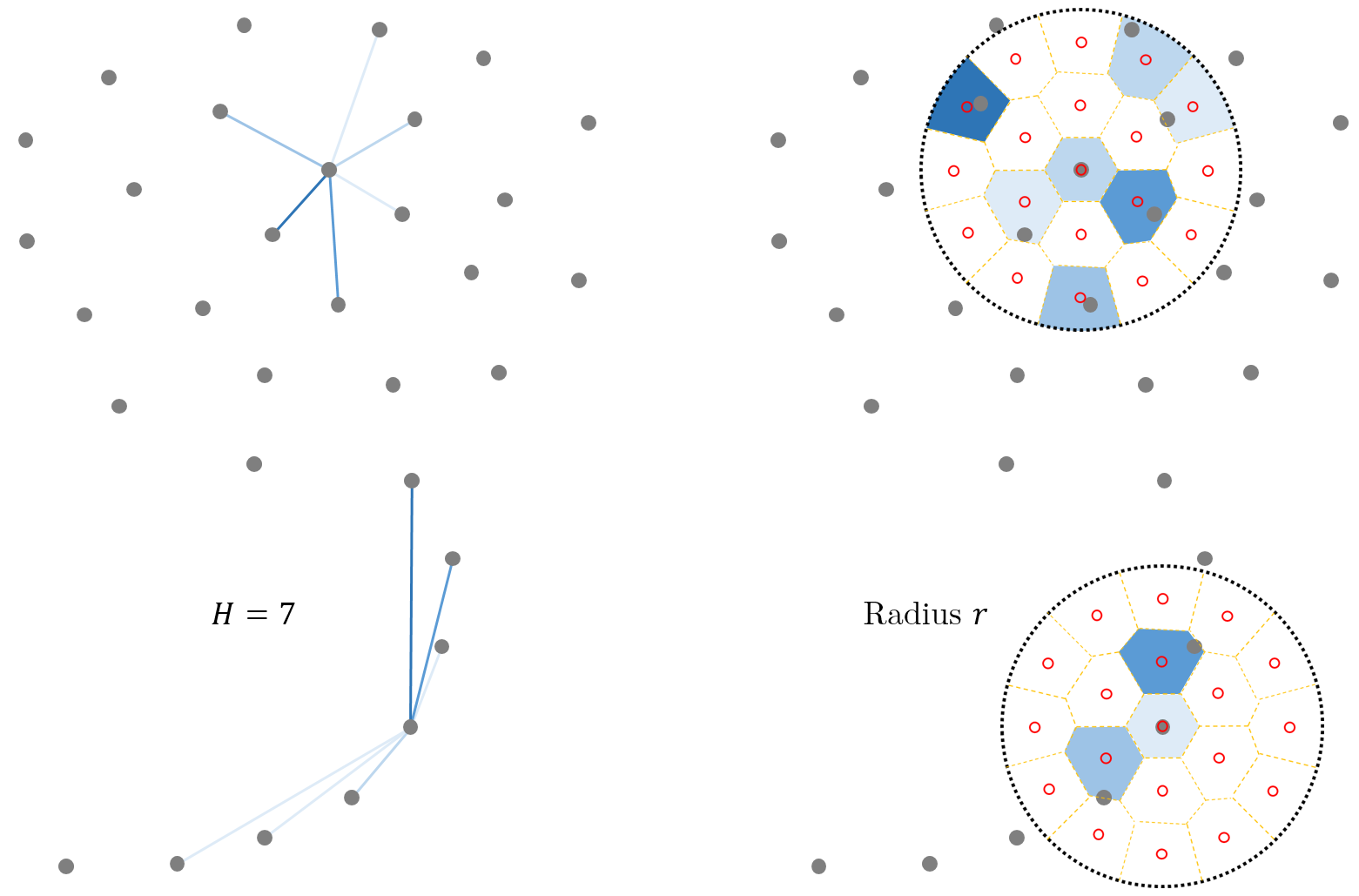}
    \caption{Illustration of our kernel attention principle, where chunks of space are weighted (right) compared to standard point self-attention where the neighbors are weighted depending on their feature instead of their position (left). In our design, no position encoding is needed, as the attention itself is a position encoding.}
    \label{fig:attention}
\end{figure}

On the contrary, the kernel attention we propose is geometric in nature. The attention weights are generated for local parts of the space, instead of being generated for neighbors depending on their features. This design gives a geometric structure to our attention mechanism, and the ability to focus on geometric patterns, as shown in \cref{fig:attention}. Therefore, our operator does not need additional position encodings to capture geometric information.

With the previous notations, we define KPConvX as

\begin{equation} \label{eq:5}
    (\mathcal{F}*g)(x) = \sum_{i<H} h_{ik^*} \left( m_{k^*} \circledcirc w_{k^*} \right) \,  \odot f_i \: ,
\end{equation}

\noindent
where $m_{k^*}$ is a vector of modulations generated for the $k^{th}$ kernel point and $\circledcirc$ is a grouped version of the Hadamard product. Indeed, $m_{k^*} \in \mathbb{R}^{C_g}$ where $C_g = C/G$ and $G$ is the number of groups. We note that with $G=C$, our design is very similar to the modulations of deformable KPConv \cite{thomas2019kpconv}, and with $G=1$, every channel gets its own modulation. In the following, we use $G=8$ unless otherwise specified. The modulations are generated all together from the feature of the central point:

\begin{equation} \label{eq:6}
    \left( m_{k} \right)_{k<K} = \alpha \left( x \right) ,
\end{equation}

\noindent
where $\alpha$ is defined as a two-layer MLP, with $C$ hidden channels, $K \times C_g$ output channels, and a sigmoid as final activation. We note that with this attention definition, augmenting the number of kernel points will affect the number of operations. However, thanks to our nearest-kernel implementation, the increase in operations is still reduced compared to the full-summation kernel.

As stated above, from this general definition we can regress to a depthwise convolution, by removing the modulations, as defined in \cref{eq:4}. But it is also possible to remove the convolution weights and only use modulations. Similarly to the image involution \cite{li2021involution}, we would then define a kernel point involution:

\begin{equation} \label{eq:7}
    (\mathcal{F} \circledast g)(x) = \sum_{i<H} h_{ik^*} \left( m_{k^*} \circledcirc f_i \right) \: .
\end{equation}

\noindent
We named this design KPInv, but it did not match the performances of KPConvX, and we do not use it in our experiments. 

KPConvX design would not work without kernel points or a similar structure. Why? The modulations are generated with $\alpha$ in a certain order, which corresponds to the order of the kernel points. If we were to produce modulations directly for neighbors, as a naive re-implementation of the image involution for point clouds would do, they would be applied in this order to the neighbors. However, even if they are ordered by coordinates or distance to the center, the neighbors' order remains highly unstable, and therefore the modulations would be applied randomly to different neighbors every time. This instability undermines the network's ability to extract geometric patterns and we observed weaker performances in the few tests we ran. 

Furthermore, predicting weights from the central point forces the network to include contextual information in its features. This contextual information is necessary for deciding where to focus attention spatially in ensuing layers. We believe this is one of the crucial properties that make KPConvX a better descriptor.

%
%
%
%

\subsection{Modern Architecture and Training}
\label{subsec:architectures}

In this work, we also focused on the design of new network architectures. We follow the path of other works that advocate the use of modern techniques for deep networks in images \cite{liu2022convnet} or point clouds \cite{qian_pointnext_2022}. Firstly, we define a small (S) and large (L) architectures, with respectively $\left[2, 2, 2, 8, 2\right]$ blocks per layer and $\left[3, 3, 9, 12, 3\right]$ blocks per layer. We choose an initial width of $64$ channels for both networks as shown in \cref{fig:network}.

For the stem of the network, we choose a full standard KPConv, to extract strong initial features as advocated in \cite{lai_stratified_2022}. The cost of this layer is negligible compared to the rest of the network.

Each network block is designed as an inverted bottleneck similarly to \cite{liu2022convnet, qian_pointnext_2022}. Compared to the standard ResNet bottleneck design \cite{he2016deep} that starts with a downsampling MLP, follows with a local extractor, and finishes with an upsampling MLP, the inverted design places the local extractor at the beginning of the block and has two MLPs that upsample and then downsample the features. We use a normalization layer (Batch Normalization) and an activation layer (Leaky ReLU), after KPConvX and after each MLP. For the final MLP of the block, the activation layer is moved after the addition.

Each layer has its own fixed number of neighbors for the local operator. In all our experiments, we use $\left[12, 16, 20, 20, 20\right]$ neighbors for each layer. The number of neighbors is lower in the early layers because the point clouds are more sparse before being subsampled.

Instead of using heavy operations for the pooling layers, we follow \cite{wu2022point} and implement partition-based pooling with a grid of increasing size. In the same fashion, we use a ratio of $2.2$ for the grid scaling between layers, which reduces the overall memory cost of the network. The grid upsampling operation projects the features back to the points that were in the same grid cell, without any modification. We also append an additional block in each decoder layer to improve the performance.

Furthermore, a common network regularization scheme, stochastic depth \cite{huang2016deep}, is implemented with DropPath operations. It consists of randomly skipping network blocks for entire elements of a batch. The implementation of DropPath for our network is not trivial because we use variable batch size as in the original KPConv \cite{thomas2019kpconv}. We refer to the supplementary material and our open-source implementation for more details.

Interestingly, we notice that the number of channels in the deeper layers does not need to be doubled to maintain good performances. Instead of keeping a ratio of $2$ between the widths of two consecutive layers, we use a ratio of $\sqrt{2}$. This reduces the size and memory consumption of our network, without affecting its performance.

\textbf{Segmentation Architecture:} After the last upsampling layer, we use a standard segmentation head, which consists of a two-layer MLP with $64$ hidden channels, and $n_\mathrm{class}$ output channels, followed by a softmax layer. At training, we use a standard cross-entropy loss.

\textbf{Classification Architecture:} The classification architecture only uses the encoder part of the network. The features are aggregated with a global average pooling and processed by the classification head, which consists of a two-layer MLP with $256$ hidden channels, and $n_\mathrm{class}$ output channels, followed by a softmax layer. We use a cross-entropy loss with label smoothing \cite{szegedy2016rethinking} for training.

Finally, we train our networks with the more recent AdamW optimizer \cite{loshchilov2017decoupled} and we use up-to-date data augmentation strategies \cite{qian_pointnext_2022}. More details and other training parameters can be found in the supplementary material.

\section{Experiments}
\label{sec:exp}

To assess the effectiveness of the proposed method, we perform experimental evaluations on three datasets: Stanford Large-Scale 3D Indoor Spaces (S3DIS) \cite{armeni20163d}, ScanNetv2 \cite{dai2017scannet}, and ScanObjectNN \cite{uy2019revisiting}. We begin by presenting our setup and comparing our results to the state of the art on these datasets. Next, we conduct several ablation studies to confirm the effectiveness of our contributions.

\begin{table}[t]
\caption{Semantic segmentation results on S3DIS Area 5 evaluated under mIoU (\%), mAcc (\%), and OA (\%) metrics.}
\vspace{-3ex}
\setlength\tabcolsep{0.5pt}
\begin{center}
\begin{tabular}{ L{3.7cm}  C{1.5cm} C{1.5cm} C{1.5cm}}

\Xhline{2\arrayrulewidth}

Method	 & mIoU	 & mAcc	 & OA	\TBstrut\\
\Xhline{2\arrayrulewidth}

KPConv \cite{thomas2019kpconv}	 & $67.1$	 & $72.8$	 & -	\\
PTv1 \cite{zhao_point_2021}	 & $70.4$	 & $76.5$	 & $90.8$	\\
SPoTr \cite{park2023self}	 & $70.8$	 & $76.4$	 & $90.7$	\\
PointNeXt \cite{qian_pointnext_2022}	 & $70.5$	 & $76.8$	 & $90.6$	\\
PointMixer \cite{choe_pointmixer_2022}	 & $71.4$	 & $77.4$	 & -	\\
PTv2 \cite{wu2022point}	 & $71.6$	 & $77.9$	 & $91.1$	\\
StratTrans \cite{lai_stratified_2022}	 & $72.0$	 & $78.1$	 & $91.5$	\\
PointVector \cite{deng2023pointvector}	 & $72.3$	 & $78.1$	 & $91.0$	\\
PointMetaBase \cite{lin2023meta}	 & $72.3$	 & -	 & $91.3$	\Bstrut\\ \hline
KPConvX-L (ours)	 & $\mathbf{73.5}$	 & $\mathbf{78.7}$	 & $\mathbf{91.7}$	\TBstrut\\

\Xhline{2\arrayrulewidth}

\end{tabular}
\end{center}
\label{table:rooms}
\vspace{-3ex}
\end{table}

%
%
%
%

\subsection{Semantic Segmentation}

\noindent
\textbf{Data and metrics.} S3DIS \cite{armeni20163d} is a challenging benchmark that consists of 6 large-scale indoor areas, with a total of 271 rooms spread across three different buildings. The points are densely sampled on the mesh surfaces and annotated with 13 semantic categories. Scannetv2 \cite{dai2017scannet}, on the other hand, is relatively larger and includes 1,201 indoor RGB-D scenes for training, 312 scenes for validation, and 100 scenes for testing. Semantic labels in 20 categories are annotated.

We adopt the standard experimental setup \cite{tchapmi2017segcloud, qi2017pointnet++, thomas2019kpconv, zhao_point_2021}, using the fifth area as our test set for S3DIS and following the official train/evaluation split for ScanNetv2. The standard metric for these datasets is the mean class-wise intersection-over-union (mIoU). Additionally, we may also present the mean of class-wise accuracy (mAcc) and the overall point-wise accuracy (OA) for S3DIS.

\begin{figure*}[b!]
    \centering
    \includegraphics[width=0.98\textwidth, keepaspectratio=true]{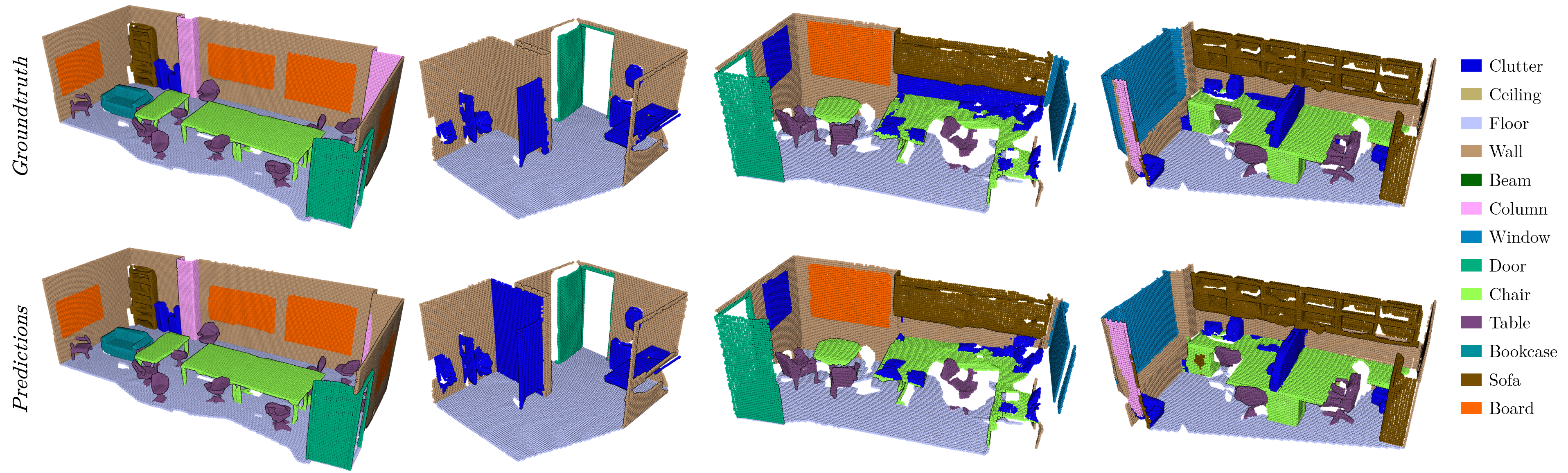}
    \caption{Visualization of Semantic segmentation results on in rooms of S3DIS Area 5.}
    \label{fig:visuS3DIS}
\end{figure*}

\noindent
\textbf{Experimental setup.} For both datasets, during training, we randomly select a point from a random room as the center of a sphere. We then sample all the points within a radius of $2.5$m to create an input point cloud. Our 5-channel input feature includes a constant one feature, RGB values, and the $z$-coordinate of the points (before augmentation) for S3DIS. For ScanNetv2, we add 3 values for the normals. Following the approach in \cite{thomas2019kpconv}, we stack these variable-length point clouds to create stacked batches. We use a target batch size of $4$ and accumulate $6$ forward passes before propagating the gradient backward. This effectively gives us a batch size of $24$, while keeping the memory consumption of a batch size of $4$.

We subsample the input point clouds using a voxel size of $0.04$m for S3DIS and $0.02$m for ScanNetv2. Similar to recent methods, we do not use spheres on the test set. Instead, we evaluate our networks on the entire rooms. More details on the training parameters and augmentations can be found in the supplementary material.

\begin{table}[t]
\caption{Semantic segmentation results on ScanNetv2 validation set evaluated under mIoU (\%) and number of parameters.}
\vspace{-3ex}
\setlength\tabcolsep{0.5pt}
\begin{center}
\begin{tabular}{ L{3.7cm}  C{2.0cm} C{2.0cm}}

\Xhline{2\arrayrulewidth}

Method	 & params	 & mIoU	\TBstrut\\
\Xhline{2\arrayrulewidth}

KPConv \cite{thomas2019kpconv}	 & $15.0$M	 & $69.2$	\Tstrut\\
PTv1 \cite{zhao_point_2021}	 & $7.8$M	 & $70.6$	\\
FastPtTrans \cite{park_fast_2022}	 & $37.9$M	 & $72.1$	\\
MinkUNet \cite{choy20194d}	 & $37.9$M	 & $72.2$	\\
LargeKernel \cite{chen2023largekernel3d}	 & $40.2$M	 & $73.2$	\\
BPNet \cite{hu2021bidirectional}	 & -	 & $73.9$	\\
StratTrans \cite{lai_stratified_2022}	 & $18.8$M	 & $74.3$	\\
EQ-Net \cite{yang2022unified}	 & -	 & $75.3$	\\
PTv2 \cite{wu2022point}	 & $11.3$M	 & $75.4$	\\
OctFormer \cite{wang2023octformer}	 & $39.0$M	 & $75.7$	\Bstrut\\ \hline
KPConvX (ours)	 & $13.5$M	 & $\mathbf{76.3}$	\TBstrut\\

\Xhline{2\arrayrulewidth}

\end{tabular}
\end{center}
\label{table:rooms}
\vspace{-3ex}
\end{table}

In contrast to \cite{qian_pointnext_2022}, we advocate for the use of voting when evaluating network predictions. We argue that if all the votes come from the same network with fixed weights, it cannot be considered an ensemble method. Moreover, when a network is trained with data augmentation, its results may vary depending on how it encounters the test data. For instance, as shown in \cite{lai_stratified_2022}, the results can unexpectedly be high when using a certain room orientation. By utilizing voting, we reduce the variance in the results and provide smoother predictions.

\noindent
\textbf{Results.} We compare our best network performance to state-of-the-art results using different versions of the dataset. The results are summarized in \cref{table:rooms}. Our KPConvX-L model achieves the highest mIoU score in both cases, outperforming the current state of the art by $\mathbf{+1.2}$ mIoU on S3DIS and by $\mathbf{+0.6}$ mIoU on Scannet. We also provide qualitative results of point cloud semantic segmentation in \cref{fig:visuS3DIS}.

%
%
%
%

\subsection{Shape Classification}

\noindent
\textbf{Data and metrics.} ScanObjectNN is a recent classification benchmark that includes approximately 15,000 real scanned objects and 2,902 unique object instances \cite{uy2019revisiting}. Due to occlusions and noise, it is considered more challenging than other 3D shape classification benchmarks. Following the standard protocol \cite{yu_point-bert_2022, qian_pointnext_2022}, we utilize the hardest variant of the dataset, PB\_T50\_RS. The performances are evaluated without pre-training using the mean of class-wise accuracy (mAcc) and overall point-wise accuracy (OA).

\noindent
\textbf{Experimental setup.} We adopt the same point resampling strategy as \cite{yu_point-bert_2022, qian_pointnext_2022}: during training, we randomly sample 1,024 points for each shape, while during testing, points are uniformly sampled. Although the input points are not subsampled with a grid, we still need to choose a fake initial grid size (0.02cm) to scale our convolution radius and define the next layer grids. Our batch size is set to 32 point clouds, with an accumulation of 2 forward passes, resulting in an effective batch size of 64. Similarly, we employ voting during testing due to the aforementioned reasons.

\noindent
\textbf{Results.} \cref{table:classif} displays our network performances compared to other competing methods. We present both the average and best scores from ten runs. Our top-performing network, KPConvX-L, outperforms PointVector \cite{deng2023pointvector} by a significant margin ($1.1\%$ mAcc and OA on average). It is important to note that we did not compare against methods that employ a pre-training strategy.

\begin{table}[t]
\caption{Shape classification on ScanObjectNN \textbf{without} pre-training. We report our average and best scores after retraining the networks ten times.}
\vspace{-3ex}
\setlength\tabcolsep{0.5pt}
\begin{footnotesize}
\begin{center}
\begin{tabular}{ L{2.2cm}  C{1.7cm} C{1.2cm} C{1.7cm} C{1.2cm}}

\Xhline{2\arrayrulewidth}
\multicolumn{1}{c}{} & \multicolumn{2}{c}{OA} & \multicolumn{2}{c}{mAcc} \TBstrut\\
\hline
Method	 & mean$\pm$std	 & best	 & mean$\pm$std	 & best	\TBstrut\\
\Xhline{2\arrayrulewidth}

PointNet \cite{qi2017pointnet}	 & -	 	 & $68.2$	 & -	 	 & $63.4$	\Tstrut\\
PointNet++ \cite{qi2017pointnet++}	 & -	 	 & $77.9$	 & -	 	 & $75.4$	\\
DGCNN \cite{wang2019dynamic}	 & -	 	 & $78.1$	 & -	 	 & $73.6$	\\
PointCNN \cite{li2018pointcnn}	 & -	 	 & $78.5$	 & -	 	 & $75.1$	\\
Simpleview \cite{goyal2021revisiting}	 & -	 	 & $80.5$	 & -	 	 & -	\\
PointBERT \cite{yu_point-bert_2022}	 & -	 	 & $83.1$	 & -	 	 & -	\\
Point-MAE \cite{pang_masked_2022}	 & -	 	 & $85.2$	 & -	 	 & -	\\
PointMLP \cite{ma_rethinking_2022}	 & $85.4	\pm0.3$	 & -	 & $83.9	\pm0.5$	 & -	\\
PointStack \cite{wijaya_advanced_2022}	 & $86.9	\pm0.3$	 & $87.2$	 & $85.8	\pm0.3$	 & $86.2$	\\
PointNeXt \cite{qian_pointnext_2022}	 & $87.7	\pm0.4$	 & $88.2$	 & $85.8	\pm0.6$	 & $86.6$	\\
PointVector \cite{deng2023pointvector}	 & $87.8	\pm0.4$	 & $88.6$	 & $86.2	\pm0.5$	 & $86.8$	\\
SPoTr \cite{park2023self}	 & -	 	 & $88.6$	 & -	 	 & $86.8$	\Bstrut\\ \hline
KPNeXt-S (ours)	 & $88.3	\pm0.4$	 & $\underline{89.0}$	 & $86.7	\pm0.5$	 & $87.4$	\Tstrut\\
KPNeXt-L (ours)	 & $\mathbf{88.9}	\pm0.3$	 & $\mathbf{89.3}$	 & $\mathbf{87.3}	\pm0.5$	 & $\mathbf{88.1}$	\Bstrut\\

\Xhline{2\arrayrulewidth}
\end{tabular}
\end{center}
\end{footnotesize}
\label{table:classif}
\vspace{-3ex}
\end{table}

%
%
%
%

\subsection{Ablation: From KPConv to KPConvX}






\begin{table}[b]
\caption{Ablation study from KPConv to KPConvX. Best mIoU is highlighted in \textbf{bold} and mIoUs within $1\%$ of the best one are \underline{underlined}.}
\vspace{-3ex}
\setlength\tabcolsep{0.5pt}
\begin{footnotesize}
\begin{center}
\begin{tabular}{ L{3.3cm}  C{1.9cm} C{1.0cm} C{0.8cm} C{0.9cm} C{1.2cm} }

\Xhline{2\arrayrulewidth}

\multicolumn{1}{c}{} & \multicolumn{1}{c}{mIoU (10 tries)}	 & TP	 & GPU	 & 	params \TBstrut\\ 
\hline
Method	 & mean$\pm$std	 & ins/s	 & GB	 & M	\TBstrut\\
\hline

KPConv (our impl)	 & $68.5	\pm0.2$	 & $35.4$	 & $12.7$	 & $14.1$	\Tstrut\\
$+$ depthwise	 & $68.0	\pm0.3$	 & $38.3$	 & $7.1$	 & $7.3$	\\
$\searrow$ radius \& neighbors	 & $68.2	\pm0.3$	 & $51.4$	 & $4.9$	 & $7.3$	\\
$+$ shell (1 14 28)	 & $69.0	\pm0.6$	 & $41.8$	 & $11.2$	 & $7.4$	\\
$+$ new architecture	 & $70.9	\pm0.7$	 & $36.6$	 & $10.1$	 & $10.1$	\\
$+$ inv-blocks	 & $\underline{72.2}	\pm0.4$	 & $41.3$	 & $8.1$	 & $7.8$	\\
$+$ nearest influence	 & $\underline{72.3}	\pm0.4$	 & $46.5$	 & $5.2$	 & $7.8$	\\
$+$ shareKP (\footnotesize{KPConvD-L})	 & $\underline{72.2}	\pm0.7$	 & $64.1$	 & $4.6$	 & $7.8$	\\
$+$ attention (\footnotesize{KPConvX-L})	 & $\mathbf{72.4}	\pm0.9$	 & $47.7$	 & $6.8$	 & $13.5$	\Bstrut\\

\Xhline{2\arrayrulewidth}
\end{tabular}
\end{center}
\end{footnotesize}
\label{table:abl0}
\vspace{-3ex}
\end{table}

In this experiment, we applied a complete series of changes to transform KPConv into KPConvX. We conducted the experiment on S3DIS dataset under controlled settings, using the same training parameters and data augmentation as in our other experiments. This is why our version of KPConv achieved a slightly better score than the original paper. In the previous experiments, we provided the results of our best model and plan to share these best weights. However, in all the following ablation studies, for each version, we trained and tested 10 identical models and got an average score. This is more reliable to provide fair comparison studies. \cref{table:abl0} shows the average and standard deviation of the 10 test mIoUs on S3DIS Area5, in addition to the best mIoU obtained. For each version, we also provide the number of million parameters, the network throughput (TP) in instances per second (ins/s), and the GPU memory consumption in GB during inference. This allows a better understanding of the impact of each change. For comparison purposes, the throughput and GPU consumption are measured using the same parameters as \cite{qian_pointnext_2022}: batches of $16$ input point clouds, each containing $15,000$ points, were fed to the network. Each model was trained and tested on a single Nvidia Tesla V100 32GB GPU.

First, we transform KPConv into a depthwise operation, reduce the radius, reduce the number of neighbors to fixed values, and add one kernel shell ($\left[1, 14\right]$ to $\left[1, 14, 28\right]$). This results in a $+0.5$ mIoU improvement.

Next, we add more layers from KP-FCNN $\left[2, 2, 2, 2, 2\right]$ to our large architecture $\left[3, 3, 9, 12, 3\right]$. We also reduce the channel scaling to $\sqrt{2}$ and add decoder layers and DropPath. With these changes, the mIoU improves by $1.9\%$, while maintaining similar GPU memory consumption and throughput. To finish the architecture changes, we use inverted bottleneck blocks for an additional $1.9\%$ mIoU gain.

We then demonstrate the effectiveness of our nearest influence design, which is $13\%$ faster (TP$\nearrow$) and $56\%$ lighter (GPU$\searrow$). Additionally, our kernel-point-sharing strategy makes the model $38\%$ faster (TP$\nearrow$) and $13\%$ lighter (GPU$\searrow$) without impacting the results.

At this point, we have our \textcolor{green1}{\textbf{KPConvD-L}} architecture, which improves performances of the original KPConv by \textcolor{green1}{\textbf{$\mathbf{+3.7}$ mIoU}} on average, runs \textcolor{green1}{\textbf{$\mathbf{80\%}$ faster}}, and uses only \textcolor{green1}{\textbf{$\mathbf{20\%}$ GPU memory}} compared to the original KPConv.

Finally, we add our attention mechanism to create the \textcolor{green2}{\textbf{KPConvX-L}} architecture. Compared to the original KPConv, it improves performances by \textcolor{green2}{\textbf{$\mathbf{+3.9}$ mIoU}} on average, runs \textcolor{green2}{\textbf{$\mathbf{35\%}$ faster}}, and uses only \textcolor{green2}{\textbf{half GPU memory}}.

%
%
%
%

\subsection{Other Ablation Studies}

\begin{table}[t]
\caption{Ablation study of the architectures. Best mIoU is highlighted in \textbf{bold} and mIoUs within $1\%$ of the best one are \underline{underlined}.}
\vspace{-3ex}
\setlength\tabcolsep{0.5pt}
\begin{footnotesize}
\begin{center}
\begin{tabular}{ L{3.0cm}  C{1.5cm} C{1.0cm} C{0.8cm}  C{1.0cm} }

\Xhline{2\arrayrulewidth}

\multicolumn{1}{c}{} & \multicolumn{1}{c}{mIoU (10-try avg)}	 & TP	 & GPU	 & params	\TBstrut\\ 
\hline
Architecture	 & mean$\pm$std	 & ins/s	 & GB	 & M	\TBstrut\\
\hline

KPConvX-L	 & $\mathbf{72.4}	\pm0.9$	 & $47.7$	 & $6.8$	 & $13.5$	\Tstrut\\
KPConvX-S	 & $71.1	\pm0.5$	 & $62.3$	 & $6.8$	 & $8.5$	\\
KPConvD-L	 & $\underline{72.2}	\pm0.7$	 & $64.1$	 & $4.6$	 & $7.8$	\\
KPConvD-S	 & $70.2	\pm0.8$	 & $75.7$	 & $4.6$	 & $5.0$	\Bstrut\\

\Xhline{2\arrayrulewidth}
\end{tabular}
\end{center}
\end{footnotesize}
\label{table:abl_arch}
\vspace{-3ex}
\end{table}

In the previous ablation study, we demonstrated the improvements brought by our contributions. In this section, we provide more insight into our method by studying the effect of some parameter changes.

First, we compare our different architectures on S3DIS Area 5. From \cref{table:abl_arch}, we observe that our best network is KPConvX-L. However, it also has the highest computational cost. Using KPConvD-L is a good option for more efficiency, this model being 35\% faster, while only sacrificing 0.2\% mIoU on average. Overall, our throughput is highly competitive and comparable to PointNeXt-XL \cite{qian_pointnext_2022}, which presented a throughput of 46 ins/s, similar to KPConvX-L in the same setup, even though their performances are only $70.5$ mIoU on average on S3DIS Area 5.

Next, we assess the impact of the number of groups in KPConvX on the performance of KPConvX-L. As shown in \cref{table:abl_groups}, when there are fewer groups, the network has a greater number of parameters to learn. Conversely, with more groups, the modulations become less effective. We observe that the network performs really well with 8 groups, which is the value we use in the rest of the paper. This value provides a balance between enhancing descriptive power and avoiding an excessive parameter burden.

Our goal is to help the reader gain a deeper understanding of the mechanisms that contribute to the performance and efficiency of our models. We thus provide more ablation and parameter studies in the supplementary material, including the convolution radius and the number of kernel points and shells.

\section{Conclusion}
\label{sec:conclusion}

We present KPConvX, an efficient feature extractor for point clouds that combines depthwise convolution and self-attention. Additionally, we introduce KPConvX-L, a new deep architecture for semantic segmentation and shape classification. KPConvX-L is trained using the latest strategies and achieves state-of-the-art performance on several benchmarks for 3D semantic segmentation and 3D shape classification.

\textbf{Limitations and future work.} Deep learning architectures utilize local feature extractors to generate new information from local neighborhoods. However, the topological or geometric nature of these local feature extractors is often understudied. Transformers and some 3D graph architectures have a topological nature, as they are based on neighbor-feature relationships. Without incorporating MLP geometric encodings, they would overlook the geometry of the point clouds. On the other hand, structured convolutions like KPConv or voxel networks are inherently geometric encodings, merging features solely based on their locations, without considering neighbor features. With KPConvX, we introduced attention into a geometric operator, in contrast to recent point transformers that incorporated geometric encodings into a topological operator. Nevertheless, in both cases, the feature extraction operators still generate new information, either topologically or geometrically. It is important to conduct thorough studies to understand how the topological or geometric nature of local feature extractors impacts the learning process, and whether these two ideas can be combined in a single architecture to separate topological and geometric features.

\begin{table}[t]
\caption{Ablation study of the KPConvX groups. Best mIoU is highlighted in \textbf{bold} and mIoUs within $1\%$ of the best one are \underline{underlined}.}
\vspace{-3ex}
\setlength\tabcolsep{0.5pt}
\begin{footnotesize}
\begin{center}
\begin{tabular}{ L{3.0cm}  C{1.5cm} C{1.0cm} C{0.8cm}  C{1.0cm} }

\Xhline{2\arrayrulewidth}

\multicolumn{1}{c}{} & \multicolumn{1}{c}{mIoU (10-try avg)}	 & TP	 & GPU	 & params	\TBstrut\\ 
\hline
Number of groups	 & mean$\pm$std	 & ins/s	 & GB	 & M	\TBstrut\\
\hline
$G = 1$	 & $\underline{72.5}	\pm0.3$	 & $32.1$	 & $9.3$	 & $47.0$	\Tstrut\\
$G = 4$	 & $\mathbf{72.5}	\pm0.4$	 & $48.2$	 & $7.2$	 & $18.3$	\\
$G = 8$*	 & $\underline{72.4}	\pm0.6$	 & $47.7$	 & $6.8$	 & $13.5$	\\
$G = 16$	 & $72.1	\pm0.5$	 & $56.8$	 & $6.7$	 & $11.1$	\\
$G = C$	 & $\underline{72.2}	\pm0.3$	 & $70.2$	 & $5.4$	 & $8.9$	\Bstrut\\

\Xhline{2\arrayrulewidth}
\end{tabular}
\end{center}
\end{footnotesize}
\label{table:abl_groups}
\vspace{-3ex}
\end{table}
{
    \small
    \bibliographystyle{ieeenat_fullname}
    \bibliography{main}
}

\appendix
\clearpage
\setcounter{page}{1}
\maketitlesupplementary

\begin{abstract}
This supplementary material is divided into the following sections.
\begin{itemize}
    \itemsep 0ex 

    \item \cref{sec:kernel} presents the kernel point initialization method.
    
    \item \cref{sec:training} details our network architectures and training parameters.

    \item \cref{sec:double} discusses a double shortcut block design.
    
    \item \cref{sec:droppath} describes how we create stochastic depth in our networks.
    
    \item \cref{sec:results} present full results on S3DIS dataset, and discusses the data preprocessing as full scenes or rooms.
    
    \item \cref{sec:moreablations} gathers additional ablation and parameter studies on S3DIS dataset.
    
    \item \cref{sec:results_scannet} present full results on Scannet dataset for our four architectures.

\end{itemize}

Our implementation can be found in the following GitHub repository:
\noindent
{\small\urlstyle{sf}\url{https://github.com/apple/ml-kpconvx}}





\end{abstract}

%
%
%
%
%
%
%
%
%

\section{Kernel Points Initialization}
\label{sec:kernel}

As explained in the main paper, we use kernel points similarly to \cite{thomas2019kpconv} and adopt a shell definition of the kernel point positions as proposed in \cite{li2021spnet}. The original KPConv \cite{thomas2019kpconv} defined the kernel points regularly on a sphere with an optimization scheme, but did not have the option to have more than one shell. SPConv
\cite{li2021spnet} proposed to initialize multiple kernels with different radii independently, and then merge them together. On the contrary, we choose to initialize all the kernels together, in a unified optimization scheme similar to the one used in KPConv but with constraints enforced on the radius of each shell.

Let $s$ be the number of shells and $\left[1, N_1, ..., N_s\right]$ be the number of points per shell. Note that we do not count the center point as a shell as it will always be alone and centered in the sphere. The first step in our kernel initialization method is to compute the shell radii $r_j$ ($j \leq s$). We distribute them regularly along the radius of the kernel sphere $r$, as shown in Fig. 3 of the main paper:

\begin{equation} \label{eq:sup1}
    \forall j \in \mathbb{N}, \quad 1 \leq j \leq s, \quad r_j = \frac{2j}{2s + 1}r \: .
\end{equation}

\noindent
For each kernel point $\widetilde{x}_k$, we apply the same repulsive potential as in \cite{thomas2019kpconv}:

\begin{equation} \label{eq:sup2}
    \forall x \in \mathbb{R}^3,\quad E_k^{rep}(x)= \frac{1}{\left\Vert x-\widetilde{x}_k \right\Vert} \: ,
\end{equation}

\noindent
but without any attractive potential. Therefore we are trying to minimize 

\begin{equation} \label{sup3}
    E^{tot} = \sum_{k<K}   \sum_{l\neq k} E_k^{rep}(\widetilde{x}_l) \: .
\end{equation}

\noindent
However, we enforce the constraint that every point can only move on the sphere defined by its shell radius:

\begin{figure}[b]
    \centering
    \includegraphics[width=0.98\columnwidth, keepaspectratio=true]{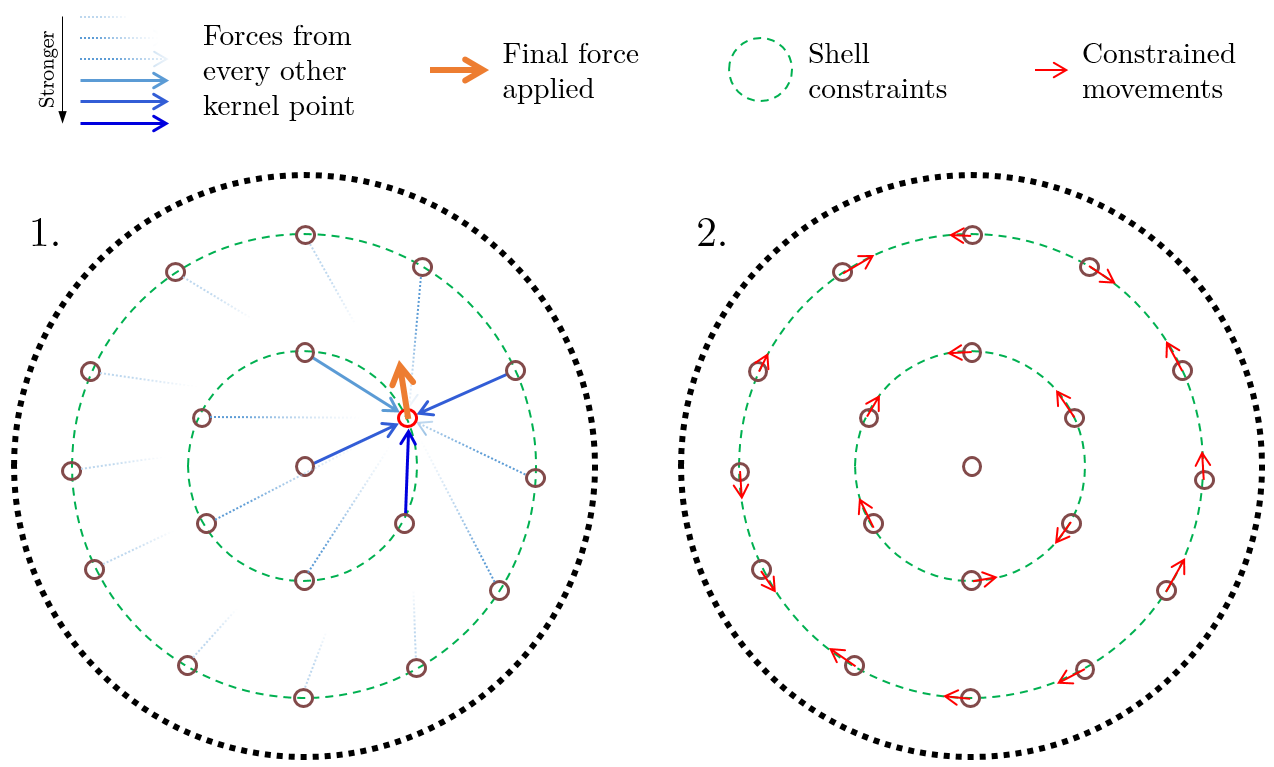}
    \caption{Illustration of our optimization function for kernel point disposition. First, the force applied to each kernel point is computed (1). Then the resulting movements are constrained to the shell spheres (2).
}
    \label{fig:supkernel}
\end{figure}

\begin{equation} \label{eq:sup4}
    \forall j \leq s, \quad K_j \leq k < K_{j+1} , \quad \left\Vert\widetilde{x}_k\right\Vert = r_j \: ,
\end{equation}

\noindent
where $K_j = 1 + N_1 + ... + N_{j-1}$. We use the same gradient descent algorithm as \cite{thomas2019kpconv}, therefore, the constraint can be applied directly to the gradients, similarly to the other constraints that were already in place, such as the one fixing the first point at the center of the sphere. The whole process is illustrated in \cref{fig:supkernel}

%
%
%
%
%
%
%
%
%

\section{Training Parameters and Augmentations}
\label{sec:training}

This section details all the training parameters and the augmentation used for our experiments.

\noindent
\textbf{S3DIS and ScanNetv2.} We share the same parameters for all our architectures. We use the following number of channels for each layer: $\left[64, 96, 128, 192, 256\right]$. As explained in the main paper, we start with $64$ features and expand with a ratio of $\sqrt{2}$, while still ensuring that the number of channels remains divisible by 16. We train for 180 epochs with 300 steps per epoch. With an effective batch size of 24, we thus see approximately 7200 input point clouds per epoch. We use an initial learning rate of $5e^{-3}$ and reduce the learning rate exponentially at each epoch, at a rate of 0.1 every 60 epochs. We use a weight decay of $0.01$ in AdamW, and standard values $\beta_1=0.9$, $\beta_2=0.999$, and $eps=1e^{-8}$. Concerning augmentations, we follow \cite{qian_pointnext_2022} with some modifications and use (in this order):
\begin{itemize}
    \itemsep 0ex 
    \item $\mathit{RandomScale}\left(s_{min}=0.9, s_{max}=1.1\right)$
    \item $\mathit{RandomFlip}\left(axis=0, p=0.5\right)$
    \item $\mathit{RandomJitter}\left(\sigma=0.005\right)$
    \item $\mathit{RandomRotate}\left(axis=2\right)$
    \item $\mathit{ChromaticAutoContrast}\left(p=0.2\right)$
    \item $\mathit{ChromaticNormalize}\left(\right)$
    \item $\mathit{RandomDropColor}\left(p=0.2\right)$
\end{itemize}

\noindent
\textbf{ScanObjectNN.} For ScanObjectNN, we use a 7-channel input feature containing a constant one feature, the point coordinates before augmentation, and the point coordinates after augmentation. The feature channels are $\left[64, 96, 128, 192, 256\right]$, following the same $\sqrt{2}$ expansion rule. Our networks are trained for 180 epochs, with enough steps to cover all the input shapes at each epoch, given a batch size of 64. The other training parameters are the same as the ones used for S3DIS. We augment the data with:
\begin{itemize}
    \itemsep 0ex 
    \item $\mathit{UnitSphereScale}\left(R=1.0\right)$
    \item $\mathit{RandomScale}\left(s_{min}=0.9, s_{max}=1.1\right)$
    \item $\mathit{RandomFlip}\left(axis=0, p=0.5\right)$
    \item $\mathit{RandomRotate}\left(axis=2\right)$
\end{itemize}

%
%
%
%
%
%
%
%
%
%

\section{Discussion on Double Shortcut Blocks}
\label{sec:double}

The common practice for deep convolutional networks and transformers is to alternate between local feature extraction (convolution or self-attention) and linear layers. It is considered more efficient to reduce the complexity of the local feature extractor, by making it depthwise and trusting the linear layers to combine features in the channel dimension. The gain in memory consumption usually allows networks to be deeper and reach better performance \cite{liu2022convnet}. Since the original ResNet bottleneck block \cite{he2016deep} was designed, it has been the base for the development of newer blocks, with a shortcut connection to solve the vanishing gradient issue.

\begin{figure}[t]
    \centering
    \includegraphics[width=0.99\columnwidth, keepaspectratio=true]{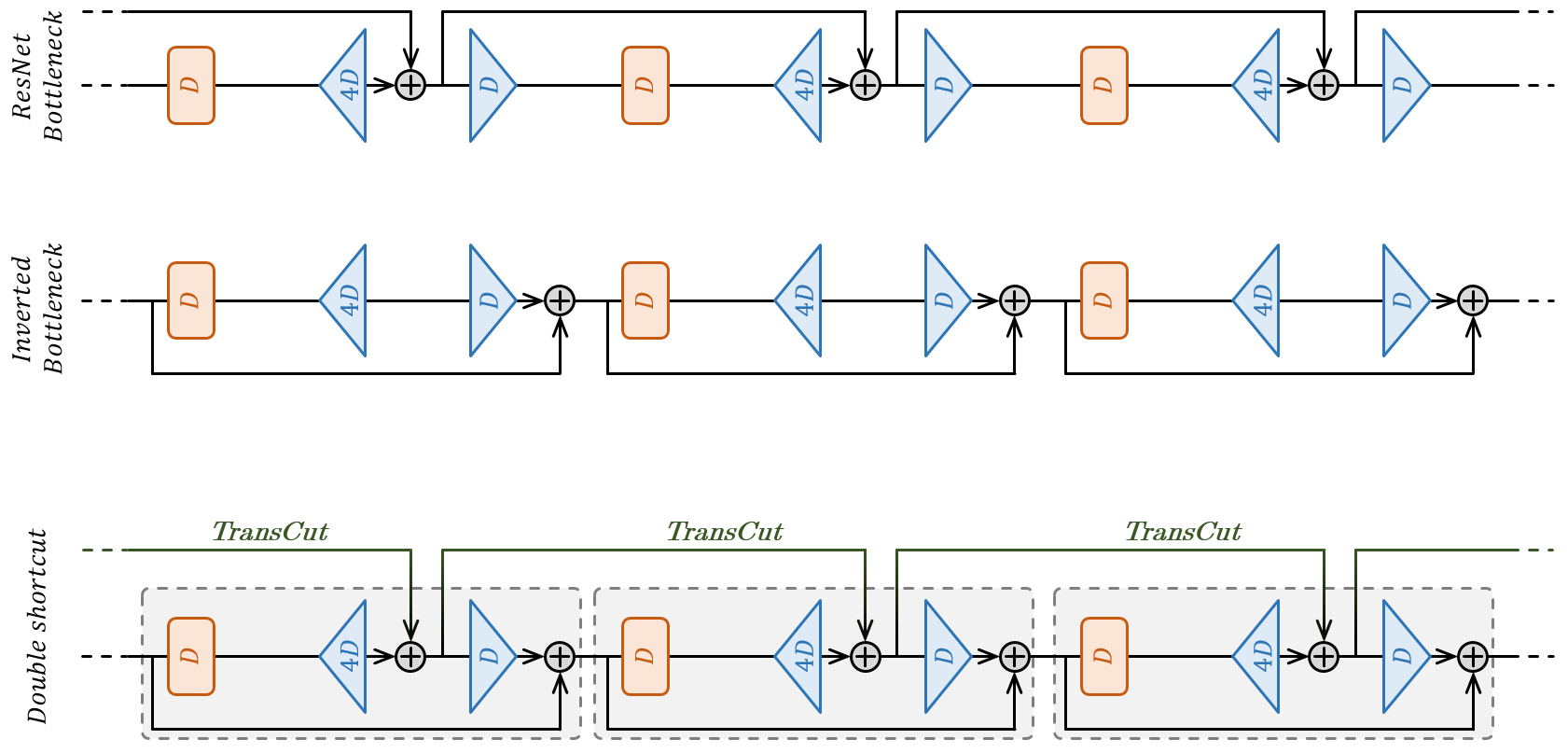}
    \caption{Illustration of our double shortcut block design, compared to ResNet bottleneck and inverted bottleneck blocks. The basic operations are the same, only the features' path changes.}
    \label{fig:blocks}
\end{figure}

More recently, the inverted bottleneck design has become more popular \cite{liu2022convnet, qian_pointnext_2022}. Compared to the bottleneck design that starts with a downsampling MLP, follows with a local extractor, and finishes with an upsampling MLP, the inverted design places the local extractor at the beginning of the block and has two MLPs that upsample and then downsample the features. Although the two designs might seem very different, they are, in fact, nearly the same. If we take a step back and look at the succession of operations, it is always the same: downsampling MLP, local extractor, upsampling MLP, downsampling MLP, etc. The only thing that changes in the series is the placement of the shortcut connection, as shown in \cref{fig:blocks}. In the first block, the shortcut propagates the high-dimensional features, and in the second, it propagates the low-dimensional features. 

During our experiments, we tested with a novel design that combines both types of shortcuts in the same computation graph. We still define blocks in the manner of the inverted bottleneck design, but we add a shortcut between consecutive blocks. This provides an additional path to help propagate the gradients for the high-dimensional features.

\begin{figure}[b]
    \centering
    \includegraphics[width=0.98\columnwidth, keepaspectratio=true]{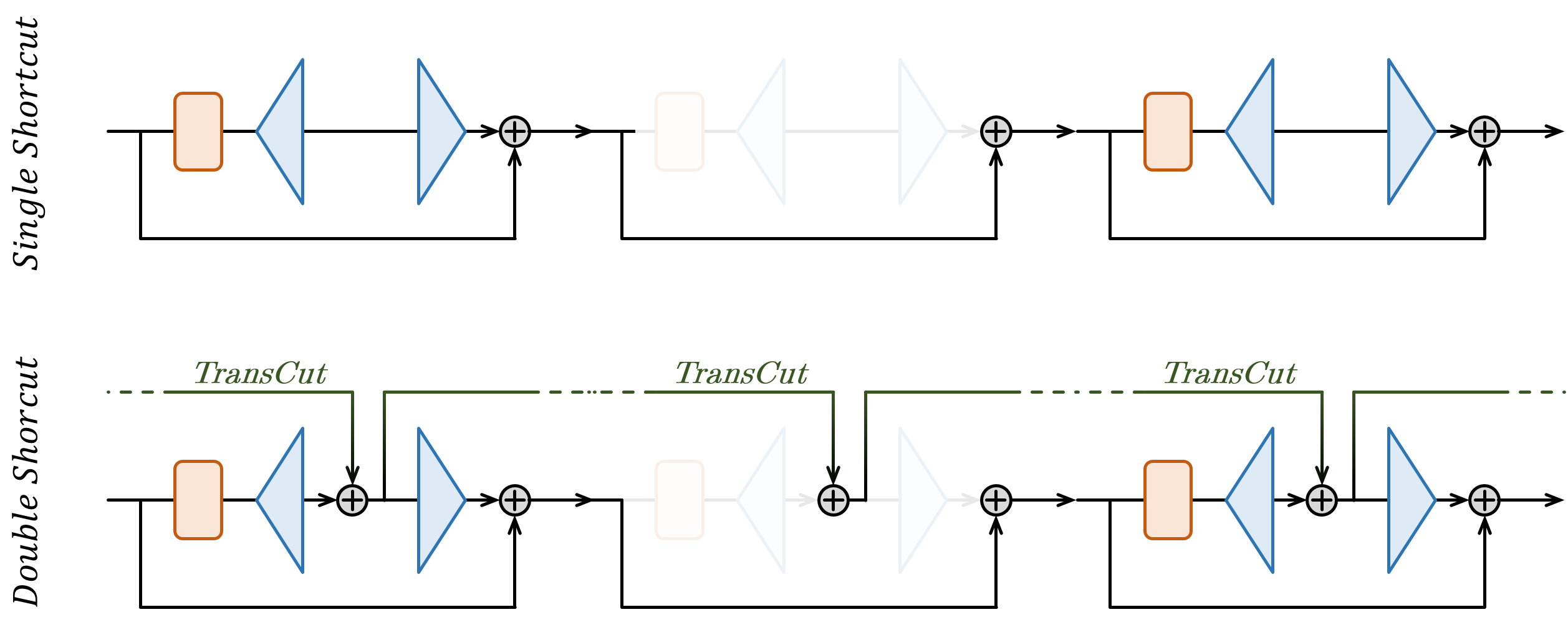}
    \caption{Stochastic depth is easily adapted to double shortcut design, by dropping the same batch elements before both shortcuts.
}
    \label{fig:supdroppath}
\end{figure}

Nevertheless, the improvement brought by the double shortcut design was not significant enough. Furthermore, additional experiments on image datasets would be necessary to validate this design, which is why we decided to keep this idea in the supplementary material.

%
%
%
%
%
%
%
%
%

\section{Stochastic Depth and DropPath Implementation}
\label{sec:droppath}

Stochastic depth \cite{huang2016deep} was proposed as a way to improve the training of deep residual networks by randomly dropping layers. For inference, all the layers are used to harness the full power of the network, allowing for better information and gradient flow. This technique highlights the significant redundancy in deep residual networks. As shown in \cref{fig:supdroppath}, the standard way to implement this is to drop (multiply by zeros) all the features from one element (point cloud) of the batch, right before the shortcut addition. This ensures that the features from the previous block are only propagated forward, as if this block did not exist. This operation is commonly referred to as DropPath.


\begin{table*}[t]
\caption{Classwise IoU for S3DIS experiment. Best results are highlighted in \textbf{bold} and results within $1\%$ of the best ones are \underline{underlined}.}
\setlength\tabcolsep{0.5pt}
\begin{footnotesize}
\begin{center}
\begin{tabular}{ L{2.3cm} | C{1.0cm} | C{0.8cm} C{0.8cm} C{0.8cm} | C{0.9cm} C{0.75cm} C{0.7cm} C{0.7cm} C{0.9cm} C{1.0cm} C{0.7cm} C{0.7cm} C{0.8cm} C{0.8cm} C{1.2cm} C{0.8cm} C{0.8cm}}
\Xhline{2\arrayrulewidth}

Method	 & Data	 & mIoU	 & mAcc	 & OA	 & ceiling	 & floor	 & wall	 & beam	 & column	 & window	 & door	 & table	 & chair	 & sofa	 & bookcase	 & board	 & clutter	\TBstrut\\ \hline

PointNet \cite{qi2017pointnet}	 & full	 & $41.1$	 & $49.0$	 & -	 & $88.8$	 & $97.3$	 & $69.8$	 & $\underline{0.1}$	 & $3.9$	 & $46.3$	 & $10.8$	 & $59.0$	 & $52.6$	 & $5.9$	 & $40.3$	 & $26.4$	 & $33.2$	\Tstrut\\
SegCloud \cite{tchapmi2017segcloud}	 & full	 & $48.9$	 & $57.4$	 & -	 & $90.1$	 & $96.1$	 & $69.9$	 & $0.0$	 & $18.4$	 & $38.4$	 & $23.1$	 & $70.4$	 & $75.9$	 & $40.9$	 & $58.4$	 & $13.0$	 & $41.6$	\\
PointCNN \cite{li2018pointcnn}	 & full	 & $57.3$	 & $63.9$	 & $85.9$	 & $92.3$	 & $98.2$	 & $79.4$	 & $0.0$	 & $17.6$	 & $22.8$	 & $62.1$	 & $74.4$	 & $80.6$	 & $31.7$	 & $66.7$	 & $62.1$	 & $56.7$	\\
SPGraph \cite{landrieu2018large}	 & full	 & $58.0$	 & $66.5$	 & $86.4$	 & $89.4$	 & $96.9$	 & $78.1$	 & $0.0$	 & $42.8$	 & $48.9$	 & $61.6$	 & $84.7$	 & $75.4$	 & $69.8$	 & $52.6$	 & $2.1$	 & $52.2$	\\
SegGCN \cite{lei2020seggcn}	 & full	 & $63.6$	 & $70.4$	 & $88.2$	 & $93.7$	 & $\underline{98.6}$	 & $80.6$	 & $0.0$	 & $28.5$	 & $42.6$	 & $74.5$	 & $88.7$	 & $80.9$	 & $71.3$	 & $69.0$	 & $44.4$	 & $54.3$	\\
MinkUNet \cite{choy20194d}	 & rooms	 & $65.4$	 & $71.7$	 & -	 & $91.8$	 & $\mathbf{98.7}$	 & $\underline{86.2}$	 & $0.0$	 & $34.1$	 & $48.9$	 & $62.4$	 & $81.6$	 & $89.8$	 & $47.2$	 & $74.9$	 & $74.4$	 & $58.6$	\\
PAConv \cite{xu_paconv_2021}	 & rooms	 & $66.6$	 & $73.0$	 & -	 & $94.6$	 & $\underline{98.6}$	 & $82.4$	 & $0.0$	 & $26.4$	 & $58.0$	 & $60.0$	 & $89.7$	 & $80.4$	 & $74.3$	 & $69.8$	 & $73.5$	 & $57.7$	\\
KPConv \cite{thomas2019kpconv}	 & full	 & $67.1$	 & $72.8$	 & -	 & $92.8$	 & $97.3$	 & $82.4$	 & $0.0$	 & $23.9$	 & $58.0$	 & $69.0$	 & $91.0$	 & $81.5$	 & $75.3$	 & $75.4$	 & $66.7$	 & $58.9$	\\
PTv1 \cite{zhao_point_2021}	 & rooms	 & $70.4$	 & $76.5$	 & $90.8$	 & $94.0$	 & $\underline{98.5}$	 & $\mathbf{86.3}$	 & $0.0$	 & $38.0$	 & $\mathbf{63.4}$	 & $74.3$	 & $89.1$	 & $82.4$	 & $74.3$	 & $\mathbf{80.2}$	 & $76.0$	 & $59.3$	\\
SPoTr \cite{park2023self}	 & rooms	 & $70.8$	 & $76.4$	 & $90.7$	 & -	 & -	 & -	 & -	 & -	 & -	 & -	 & -	 & -	 & -	 & -	 & -	 & -	\\
PointNeXt \cite{qian_pointnext_2022}	 & rooms	 & $70.5$	 & $76.8$	 & $90.6$	 & $94.2$	 & $\underline{98.5}$	 & $84.4$	 & $0.0$	 & $37.7$	 & $59.3$	 & $74.0$	 & $83.1$	 & $91.6$	 & $77.4$	 & $77.2$	 & $78.8$	 & $60.6$	\\
PointMixer \cite{choe_pointmixer_2022}	 & rooms	 & $71.4$	 & $77.4$	 & -	 & $94.2$	 & $98.2$	 & $\underline{86.0}$	 & $0.0$	 & $43.8$	 & $62.1$	 & $78.5$	 & $90.6$	 & $82.2$	 & $73.9$	 & $\underline{79.8}$	 & $78.5$	 & $59.4$	\\
PTv2 \cite{wu2022point}	 & rooms	 & $71.6$	 & $77.9$	 & $91.1$	 & -	 & -	 & -	 & -	 & -	 & -	 & -	 & -	 & -	 & -	 & -	 & -	 & -	\\
StratTrans \cite{lai_stratified_2022}	 & rooms	 & $72.0$	 & $78.1$	 & $\underline{91.5}$	 & $\mathbf{96.2}$	 & $\mathbf{98.7}$	 & $85.6$	 & $0.0$	 & $\mathbf{46.1}$	 & $60.0$	 & $76.8$	 & $\mathbf{92.6}$	 & $84.5$	 & $77.8$	 & $75.2$	 & $78.1$	 & $\mathbf{64.0}$	\\
PointVector \cite{deng2023pointvector}	 & rooms	 & $72.3$	 & $78.1$	 & $91.0$	 & $95.1$	 & $\underline{98.6}$	 & $85.1$	 & $0.0$	 & $41.4$	 & $60.8$	 & $76.7$	 & $84.4$	 & $\underline{92.1}$	 & $82.0$	 & $77.2$	 & $85.1$	 & $61.4$	\\
PtMetaBase \cite{lin2023meta}	 & rooms	 & $72.3$	 & -	 & $\underline{91.3}$	 & -	 & -	 & -	 & -	 & -	 & -	 & -	 & -	 & -	 & -	 & -	 & -	 & -	\Bstrut\\ \hline
KPConvX-L (ours)	 & rooms	 & $\mathbf{73.5}$	 & $\mathbf{78.7}$	 & $\mathbf{91.7}$	 & $94.9$	 & $\underline{98.5}$	 & $\underline{86.2}$	 & $\mathbf{0.1}$	 & $40.4$	 & $63.0$	 & $\mathbf{84.1}$	 & $84.0$	 & $\mathbf{92.4}$	 & $\mathbf{82.5}$	 & $79.0$	 & $\mathbf{86.8}$	 & $63.1$	\TBstrut\\

\Xhline{2\arrayrulewidth}
\end{tabular}
\end{center}
\end{footnotesize}
\label{table:supS3DIS}
\end{table*}

\begin{table}[b]
\caption{Architecture study for KPConvX. Best results are highlighted in \textbf{bold} and results within $1\%$ of the best ones are \underline{underlined}.}
\setlength\tabcolsep{0.5pt}
\begin{footnotesize}
\begin{center}
\begin{tabular}{ L{3.0cm}  C{1.5cm} C{1.0cm} C{0.8cm}  C{1.0cm} }

\Xhline{2\arrayrulewidth}

\multicolumn{1}{c}{} & \multicolumn{1}{c}{mIoU (5-try avg)}	 & TP	 & GPU	 & params	\TBstrut\\ 
\hline
Architecture	 & mean$\pm$std	 & ins/s	 & GB	 & M	\TBstrut\\
\hline

$\left[4,4,12,20,4\right]+1$	 & $\underline{72.3}	\pm0.6$	 & $38.1$	 & $6.9$	 & $19.7$	\Tstrut\\
$\left[5,5,13,21,4\right]+0$	 & $71.6	\pm0.7$	 & $42.8$	 & $4.6$	 & $19.7$	\\
$\left[4,4,12,20,4\right]+0$	 & $\underline{72.1}	\pm0.7$	 & $48.2$	 & $4.6$	 & $18.7$	\\
$\left[3,3,9,12,3\right]+1$ *	 & $\mathbf{72.4}	\pm0.9$	 & $47.7$	 & $6.8$	 & $13.5$	\\
$\left[4,4,4,12,4\right]+1$	 & $\underline{72.2}	\pm0.6$	 & $46.1$	 & $6.8$	 & $13.4$	\\
$\left[3,3,3,9,3\right]+1$	 & $\underline{72.1}	\pm0.6$	 & $52.6$	 & $6.8$	 & $10.4$	\\
$\left[2,2,2,8,2\right]+1$	 & $71.7	\pm0.4$	 & $62.3$	 & $6.8$	 & $8.5$	\\
$\left[2,2,2,6,2\right]+1$	 & $71.5	\pm0.8$	 & $64.1$	 & $6.8$	 & $7.4$	\\
$\left[2,2,2,2,2\right]+1$	 & $70.9	\pm0.5$	 & $64.3$	 & $6.8$	 & $5.2$	\\
$\left[3,3,3,3,3\right]+0$	 & $70.0	\pm0.3$	 & $75.7$	 & $4.6$	 & $6.2$	\\
$\left[2,2,2,2,2\right]+0$	 & $68.8	\pm0.5$	 & $88.7$	 & $4.6$	 & $4.2$	\Bstrut\\

\Xhline{2\arrayrulewidth}
\end{tabular}
\end{center}
\end{footnotesize}
\label{table:abl_archX}
\end{table}

\begin{table}[b]
\caption{Architecture study for KPConvD. Best results are highlighted in \textbf{bold} and results within $1\%$ of the best ones are \underline{underlined}.}
\setlength\tabcolsep{0.5pt}
\begin{footnotesize}
\begin{center}
\begin{tabular}{ L{3.0cm}  C{1.5cm} C{1.0cm} C{0.8cm}  C{1.0cm} }

\Xhline{2\arrayrulewidth}

\multicolumn{1}{c}{} & \multicolumn{1}{c}{mIoU (5-try avg)}	 & TP	 & GPU	 & params	\TBstrut\\ 
\hline
Architecture	 & mean$\pm$std	 & ins/s	 & GB	 & M	\TBstrut\\
\hline

$\left[4,4,12,20,4\right]+1$	 & $\underline{72.1}	\pm0.4$	 & $59.5$	 & $4.6$	 & $11.3$	\Tstrut\\
$\left[5,5,13,21,4\right]+0$	 & $71.8	\pm0.3$	 & $57.1$	 & $4.6$	 & $11.3$	\\
$\left[4,4,12,20,4\right]+0$	 & $71.4	\pm0.5$	 & $68.9$	 & $4.6$	 & $10.8$	\\
$\left[3,3,9,12,3\right]+1$ *	 & $\mathbf{72.2}	\pm0.7$	 & $64.1$	 & $4.6$	 & $7.8$	\\
$\left[4,4,4,12,4\right]+1$	 & $71.3	\pm0.8$	 & $50.5$	 & $4.6$	 & $7.8$	\\
$\left[3,3,3,9,3\right]+1$	 & $71.4	\pm0.4$	 & $65.5$	 & $4.6$	 & $6.1$	\\
$\left[2,2,2,8,2\right]+1$	 & $71.3	\pm0.5$	 & $75.7$	 & $4.6$	 & $5.0$	\\
$\left[2,2,2,6,2\right]+1$	 & $71.2	\pm0.4$	 & $79.5$	 & $4.6$	 & $4.4$	\\
$\left[2,2,2,2,2\right]+1$	 & $70.6	\pm0.6$	 & $85.7$	 & $4.6$	 & $3.2$	\\
$\left[3,3,3,3,3\right]+0$	 & $70.2	\pm0.4$	 & $94.0$	 & $4.6$	 & $3.7$	\\
$\left[2,2,2,2,2\right]+0$	 & $69.5	\pm0.3$	 & $115.5$	 & $4.6$	 & $2.6$	\Bstrut\\

\Xhline{2\arrayrulewidth}
\end{tabular}
\end{center}
\end{footnotesize}
\label{table:abl_archD}
\end{table}

In cases where all the batch elements are of the same length, implementing this technique is straightforward. However, in our case where the batch elements are stacked along the first dimension, we need to obtain a mask indicating the dropped batch elements. Furthermore, when using our double shortcut design, we need to perform the DropPath operation before both shortcuts, using the same mask for the same batch elements. We provide a custom implementation of the common DropPath operation in our open-source code, enabling future work to utilize it as well.

%
%
%
%
%
%
%
%
%

\section{Full Results on S3DIS and Discussion about Data Preprocessing}
\label{sec:results}

As mentioned in the paper, we provide the full classwise IoU of our best model on S3DIS Area5 in \cref{table:supS3DIS}. We also want to highlight an important factor in the results of this dataset that is often overlooked in previous work: there are two ways to preprocess the data of S3DIS.

On the one hand, it is possible to load entire areas as very large scenes and sample subsets (spheres or cubes) for training \cite{qi2017pointnet++, thomas2019kpconv}. On the other hand, it is also possible to load single rooms as smaller scenes and use them as input, optionally dropping some points \cite{qian_pointnext_2022, zhao_point_2021}. We observe a significant improvement when using the rooms, and we choose this strategy in this paper. For transparency, we indicate which data preprocessing (rooms or full scenes) was used for each method in the state of the art in \cref{table:supS3DIS}.

%
%
%
%
%
%
%
%
%

\section{Additional Ablation and Parameter Studies}
\label{sec:moreablations}

In this section, we provide additional ablation and parameter studies that were not crucial for the paper but are interesting for the reader to gain more insight into the mechanisms of our approach. For this larger study, we provide the average score over 5 attempts.

\begin{table}[t]
\caption{Study of the kernel point shells. Best results are highlighted in \textbf{bold} and results within $1\%$ of the best ones are \underline{underlined}.}
\setlength\tabcolsep{0.5pt}
\begin{footnotesize}
\begin{center}
\begin{tabular}{ L{3.0cm}  C{1.5cm} C{1.0cm} C{0.8cm}  C{1.0cm} }

\Xhline{2\arrayrulewidth}

\multicolumn{1}{c}{} & \multicolumn{1}{c}{mIoU (5-try avg)}	 & TP	 & GPU	 & params	\TBstrut\\ 
\hline
Kernel Point Shells	 & mean$\pm$std	 & ins/s	 & GB	 & M	\TBstrut\\
\hline

$\left[1,6\right]$	 & $70.6	\pm0.4$	 & $65.9$	 & $3.2$	 & $9.3$	\Tstrut\\
$\left[1,12\right]$	 & $70.9	\pm0.9$	 & $61.4$	 & $3.6$	 & $10.0$	\\
$\left[1,14\right]$	 & $71.4	\pm1.1$	 & $59.7$	 & $3.7$	 & $10.2$	\\
$\left[1,19\right]$	 & $71.5	\pm0.7$	 & $57.4$	 & $4.1$	 & $10.8$	\\
$\left[1,28\right]$	 & $71.8	\pm0.5$	 & $52.5$	 & $5.0$	 & $11.9$	\\
$\left[1,12,14\right]$	 & $71.7	\pm0.7$	 & $51.8$	 & $4.7$	 & $11.6$	\\
$\left[1,12,19\right]$	 & $\underline{72.3}	\pm0.5$	 & $52.4$	 & $5.4$	 & $12.2$	\\
$\left[1,12,28\right]$	 & $72.0	\pm0.4$	 & $50.0$	 & $6.6$	 & $13.3$	\\
$\left[1,14,19\right]$	 & $71.9	\pm0.9$	 & $51.3$	 & $5.7$	 & $12.5$	\\
$\left[1,14,28\right]$*	 & $\mathbf{72.4}	\pm0.6$	 & $47.7$	 & $6.8$	 & $13.5$	\\
$\left[1,14,35\right]$	 & $\underline{72.3}	\pm0.5$	 & $46.1$	 & $7.7$	 & $14.3$	\\
$\left[1,14,42\right]$	 & $\underline{72.3}	\pm0.7$	 & $42.9$	 & $8.6$	 & $15.1$	\\
$\left[1,19,28\right]$	 & $72.0	\pm0.5$	 & $44.5$	 & $7.5$	 & $14.1$	\\
$\left[1,19,35\right]$	 & $\underline{72.2}	\pm0.5$	 & $45.8$	 & $8.4$	 & $14.9$	\\
$\left[1,19,42\right]$	 & $\underline{72.4}	\pm0.3$	 & $42.2$	 & $9.3$	 & $15.7$	\Bstrut\\

\Xhline{2\arrayrulewidth}
\end{tabular}
\end{center}
\end{footnotesize}
\label{table:abl_shells}
\end{table}

First, we present full architecture studies for KPConvX and KPConvD, respectively. For the purpose of this experiment, we define our architectures as $\left[N_1, N_2, N_3, N_4, N_5\right] + N_{dec}$, where $N_i$ is the number of blocks for layer $i$ and $N_{dec}$ is the number of decoder blocks used (same for each layer). For clarity, in the main paper, we chose to highlight two architectures: small ($\left[2, 2, 2, 8, 2\right] + 1$) and large ($\left[3, 3, 9, 12, 3\right] + 1$). In \cref{table:abl_archX} and \cref{table:abl_archD}, we can find these two architectures along with other architecture variants. The architecture sizes vary from the original one used by KPConv ($\left[2, 2, 2, 2, 2\right] + 0$) to an extremely large architecture $\left[4, 4, 12, 20, 4\right] + 1$. We find that bigger architectures perform better than smaller architectures, but when reaching an extremely large size above KPConvX-L, the performance drops again. We also notice that adding a decoder layer improves the performance even compared to an architecture that has one more encoder layer to compensate.

Then, we showcase a study of the number of kernel points and shells for KPConvX in \cref{table:abl_shells}. The studied kernel point dispositions range from a very simple one-shell $\left[1, 6\right]$ disposition where each kernel point is placed in a cardinal direction, to a large two-shell $\left[1, 19, 42\right]$ disposition. We observe a general trend where larger kernels improve the results. However, similarly to the architecture study, we observe that the performance stops improving if the number of kernel points increases too much. We thus chose the $\left[1, 14, 28\right]$ disposition which led to the best results in this experiment.

\begin{table}[t]
\caption{Parameter study of the convolution radius. Best results are highlighted in \textbf{bold} and results within $1\%$ of the best ones are 
\underline{underlined}.}
\setlength\tabcolsep{0.5pt}
\begin{footnotesize}
\begin{center}
\begin{tabular}{ L{3.0cm}  C{1.5cm} C{1.0cm} C{0.8cm}  C{1.0cm} }

\Xhline{2\arrayrulewidth}

\multicolumn{1}{c}{} & \multicolumn{1}{c}{mIoU (5-try avg)}	 & TP	 & GPU	 & params	\TBstrut\\ 
\hline
Convolution radius	 & mean$\pm$std	 & ins/s	 & GB	 & M	\TBstrut\\
\hline

$r = 1.3$	 & $71.7	\pm0.5$	 & $47.6$	 & $6.8$	 & $13.5$	\Tstrut\\
$r = 1.4$	 & $\underline{72.1}	\pm0.5$	 & $47.8$	 & $6.8$	 & $13.5$	\\
$r = 1.5$	 & $71.9	\pm0.4$	 & $47.5$	 & $6.8$	 & $13.5$	\\
$r = 1.6$	 & $71.8	\pm0.4$	 & $47.9$	 & $6.8$	 & $13.5$	\\
$r = 1.7$	 & $71.9	\pm0.3$	 & $47.8$	 & $6.8$	 & $13.5$	\\
$r = 1.8$	 & $71.9	\pm0.7$	 & $47.2$	 & $6.8$	 & $13.5$	\\
$r = 1.9$	 & $\underline{72.1}	\pm0.4$	 & $48.1$	 & $6.8$	 & $13.5$	\\
$r = 2$	 & $\underline{72.1}	\pm0.5$	 & $47.4$	 & $6.8$	 & $13.5$	\\
$r = 2.1$ *	 & $\mathbf{72.4}	\pm0.6$	 & $47.7$	 & $6.8$	 & $13.5$	\\
$r = 2.2$	 & $71.9	\pm0.4$	 & $47.1$	 & $6.8$	 & $13.5$	\\
$r = 2.3$	 & $72.0	\pm0.3$	 & $47.8$	 & $6.8$	 & $13.5$	\\
$r = 2.4$	 & $\underline{72.2}	\pm0.6$	 & $47.8$	 & $6.8$	 & $13.5$	\\
$r = 2.5$	 & $\underline{72.0}	\pm0.6$	 & $47.4$	 & $6.8$	 & $13.5$	\\
$r = 2.6$	 & $\underline{72.1}	\pm0.4$	 & $47.9$	 & $6.8$	 & $13.5$	\\
$r = 2.7$	 & $72.0	\pm0.4$	 & $47.9$	 & $6.8$	 & $13.5$	\\
$r = 2.8$	 & $71.9	\pm0.3$	 & $47.8$	 & $6.8$	 & $13.5$	\\
$r = 2.9$	 & $71.7	\pm0.5$	 & $47.2$	 & $6.8$	 & $13.5$	\\
$r = 3$	 & $72.0	\pm0.5$	 & $47.2$	 & $6.8$	 & $13.5$	\\
$r = 3.1$	 & $71.6	\pm0.5$	 & $46.7$	 & $6.8$	 & $13.5$	\Bstrut\\

\Xhline{2\arrayrulewidth}
\end{tabular}
\end{center}
\end{footnotesize}
\label{table:abl_radius}
\end{table}

Finally, we also study the radius of our convolution kernel in \cref{table:abl_radius}. This parameter does not affect the network's size or efficiency. The effect of changing the convolution radius is that it changes the position of the kernel points in space, scaling the radius of each shell accordingly. The kernel points will be associated with different neighbors depending on their position. If the radius is too small, further neighbors will not have any associated kernel points, and the kernel points placed near the center will be less likely to have any associated neighbors. If the radius is too large, the area covered by each kernel point will be bigger, and the kernel will thus be less descriptive, missing finer details in the input point patterns. Therefore, we find an optimal radius value of $2.1$. As a reminder, the radius value is defined relative to the subsampling grid size at every layer. For example, with a $2.1$ radius, the first convolution radius on S3DIS data, which is subsampled at $4$cm, is $8.4$cm.



\begin{table*}[t!]
\caption{Classwise IoU for Scannet experiment. Best results are highlighted in \textbf{bold} and results within $1\%$ of the best ones are \underline{underlined}.}
\setlength\tabcolsep{0.5pt}
\begin{footnotesize}
\begin{center}
\begin{tabular}{ L{1.5cm} | C{0.7cm} | C{0.7cm} C{0.7cm} C{0.7cm} C{0.7cm} C{0.7cm} C{0.7cm} C{0.7cm} C{0.7cm} C{0.7cm} C{0.7cm} C{0.7cm} C{0.7cm} C{0.7cm} C{0.7cm} C{0.7cm} C{0.7cm} C{0.7cm} C{0.7cm} C{0.7cm} C{0.7cm}}

\multicolumn{1}{c}{Model}	 & \rott{mIoU}	 & \rott{otherfurniture}	 & \rott{bathtub}	 & \rott{sink}	 & \rott{toilet}	 & \rott{shower}	 & \rott{refridgerator}	 & \rott{curtain}	 & \rott{desk}	 & \rott{counter}	 & \rott{picture}	 & \rott{bookshelf}	 & \rott{window}	 & \rott{door}	 & \rott{table}	 & \rott{sofa}	 & \rott{chair}	 & \rott{bed}	 & \rott{cabinet}	 & \rott{floor}	 & \rott{wall}	\TBstrut\\ \hline

KPConvX-L	 & $\mathbf{76.3}$	 & $61.9$	 & $85.6$	 & $71.5$	 & $93.9$	 & $64.3$	 & $\underline{64.5}$	 & $77.2$	 & $68.8$	 & $70.0$	 & $\mathbf{40.5}$	 & $81.8$	 & $\mathbf{74.2}$	 & $\mathbf{72.8}$	 & $\mathbf{79.6}$	 & $\mathbf{85.6}$	 & $92.2$	 & $\underline{84.4}$	 & $\mathbf{74.0}$	 & $\mathbf{95.9}$	 & $\mathbf{87.2}$	\Tstrut\\
KPConvX-S	 & $75.7$	 & $\mathbf{63.9}$	 & $88.1$	 & $70.6$	 & $\mathbf{95.0}$	 & $68.2$	 & $\mathbf{64.6}$	 & $77.9$	 & $64.1$	 & $\mathbf{71.1}$	 & $38.1$	 & $83.0$	 & $69.0$	 & $72.1$	 & $76.3$	 & $84.1$	 & $91.6$	 & $82.6$	 & $70.2$	 & $\underline{95.7}$	 & $\mathbf{87.2}$	\\
KPConvD-L	 & $\underline{76.2}$	 & $61.1$	 & $87.9$	 & $\mathbf{72.4}$	 & $94.1$	 & $\mathbf{74.0}$	 & $61.8$	 & $\mathbf{80.1}$	 & $\mathbf{69.9}$	 & $70.2$	 & $35.4$	 & $\mathbf{84.7}$	 & $69.2$	 & $70.8$	 & $77.2$	 & $84.8$	 & $\mathbf{92.8}$	 & $\mathbf{84.7}$	 & $70.5$	 & $\underline{95.7}$	 & $\underline{86.8}$	\\
KPConvD-S	 & $75.5$	 & $63.4$	 & $\mathbf{89.9}$	 & $71.3$	 & $92.2$	 & $67.2$	 & $62.1$	 & $78.5$	 & $64.5$	 & $68.4$	 & $39.7$	 & $82.6$	 & $70.5$	 & $72.0$	 & $75.3$	 & $81.2$	 & $91.5$	 & $\mathbf{84.7}$	 & $71.3$	 & $\mathbf{95.9}$	 & $\underline{87.0}$	\Bstrut\\
\Xhline{2\arrayrulewidth}
\end{tabular}
\end{center}
\end{footnotesize}
\label{table:supScannet}
\end{table*}

%
%
%
%
%
%
%
%
%

\section{Full Results on Scannet}
\label{sec:results_scannet}

We also provide the full classwise IoU for our 4 models on the Scannet validation set. As shown in \cref{table:supScannet}. KPConvX-L is our best network on this dataset as well, followed closely by KPConvD-L. Note that, as opposed to S3DIS, Scannet input point clouds can only be defined as rooms.



\end{document}